\documentclass[runningheads]{llncs}

\usepackage{eccv}

\usepackage{eccvabbrv}
\usepackage{multirow}
\usepackage{graphicx}
\usepackage{booktabs}
\usepackage{colortbl}
\usepackage{xcolor}
\usepackage{tabularx}
\usepackage{adjustbox}

\usepackage{makecell}
\usepackage{wrapfig} 
\usepackage{float}
\usepackage[table]{xcolor}
\definecolor{mygray}{gray}{0.94}
\usepackage[accsupp]{axessibility}  

\usepackage{hyperref}

\usepackage{orcidlink}
\usepackage{needspace}

\begin{document}

\title{Ring Forcing: Towards Precise Long-Term Memory for Autoregressive Video Diffusion} 

\titlerunning{Ring Forcing}
\newcommand{\name}{Ring Forcing}
\author{
Bowen Xue\inst{1} \and
Brandon Y. Feng\inst{2} \and
Chenguo Lin\inst{3} \and
Yuchen Lin\inst{3} \and
Yujia Zeng\inst{4} \and
Lvmin Zhang\inst{1} \and
Maneesh Agrawala\inst{1} \and
Honglei Yan\inst{5} \and
Panwang Pan\inst{5}\thanks{Project lead and corresponding author.}
}

\authorrunning{B.~Xue et al.}

\institute{
Stanford University, Stanford, CA, USA\\
\email{bowenxue2005@gmail.com, \{lvmin,maneesh\}@cs.stanford.edu}
\and
Massachusetts Institute of Technology, Cambridge, MA, USA\\
\email{brandon.fengys@gmail.com}
\and
Peking University, Beijing, China\\
\email{\{chenguolin,linyuchen\}@stu.pku.edu.cn}
\and
University of California, Berkeley, Berkeley, CA, USA\\
\email{yujiazng@gmail.com}
\and
ByteDance, Beijing, China\\
\email{yanhonglei@bytedance.com, paulpanwang@gmail.com}
}
\maketitle
\vspace{-4mm}
\begin{center}
    \textbf{Project Page: }
    \url{https://ringforcing.com}
\end{center}
\begin{center}
    \centering
    \captionsetup{type=figure}
    \includegraphics[width=\textwidth]{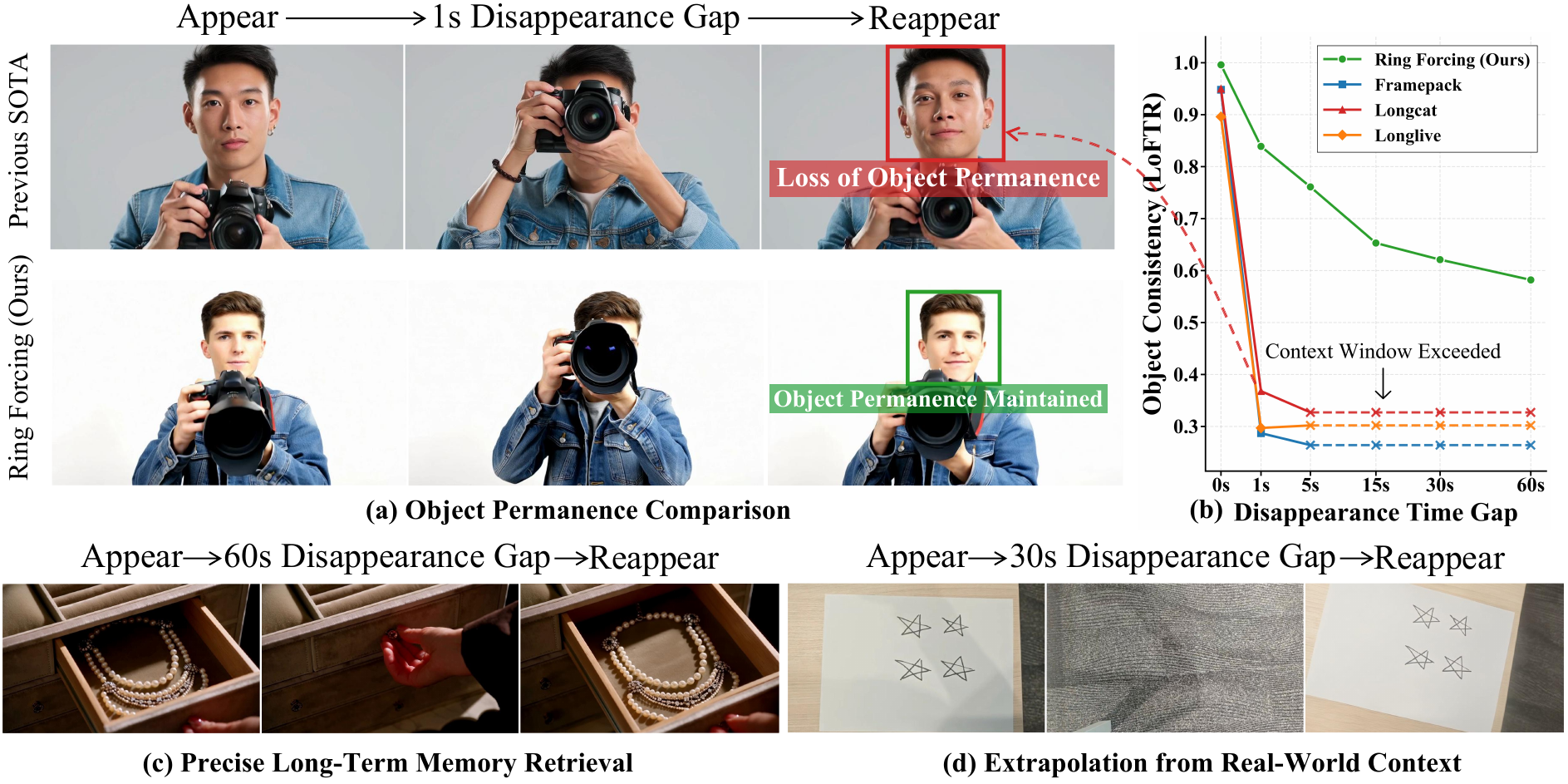}
\caption{\textbf{Ring Forcing endows autoregressive video diffusion models with precise long-term memory and robust object permanence.} 
    It preserves exact object identity through prolonged occlusions and disappearances, significantly outperforming SOTA baselines \textbf{(a, b)}. 
    Even across an extreme 60-second gap, our method faithfully recovers complex high-frequency details from distant history \textbf{(c)}. 
    This effectively overcomes the limitations of short context windows and generalizes robustly to unconstrained real-world video histories \textbf{(d)}.}
  \label{fig:teaser}
\end{center}

\begin{abstract}

Scaling video generation to long durations reveals a critical bottleneck: current models lack robust long-term memory. 
This deficiency can be studied along two critical aspects: \textit{object permanence}, the ability to precisely reproduce the appearance of objects upon re-entry; and \textit{memory capacity}, the ability to process ultra-long context and use information from distant history. 
Robust long-term memory requires both: object permanence without sufficient context handling limits the temporal scope, while long context length without permanence fails to maintain identity. 
To address this, we present \textbf{Ring Forcing}, an autoregressive video diffusion framework designed to robustly construct and precisely utilize long-term memory. 
Our ring-structured training strategy enforces retrieval from distant history, effectively reconciling the trade-off between strict historical adherence and generative diversity. 
To expand memory capacity, we introduce a compression and timestep composition strategy. 
Under fixed sequence length constraints, this method extends the effective historical span to minutes-long durations and achieves a comprehensive receptive field over the entire history.
Furthermore, we present a sparse RoPE mechanism to enable flexible, scalable memory adaptation while fully exploiting pre-trained priors. 
Extensive experiments demonstrate that \name{} achieves superior minutes-long coherence and object permanence, significantly outperforming state-of-the-art methods.

\end{abstract}


\section{Introduction}
\label{sec:intro}
Diffusion Transformers have recently advanced video generation, producing short clips with striking realism~\cite{sora2,gemini_video_generation,kling_ai_global,wan,longcat}.
However, the research frontier is no longer confined to a few seconds: emerging applications increasingly demand minutes-long rollouts with sustained narrative coherence~\cite{skyreels, magi, causvid, selfforcing}.
At this horizon, the central obstacle is not local smoothness but \emph{long-term memory}: the ability to preserve and reuse information that may disappear from view and return much later.
Autoregressive video models~\cite{selfforcing,rollingforcing,deepforcing,longcat} still struggle with \textit{object permanence}.
A canonical failure occurs when an object exits the camera frustum and later re-enters: despite being observed earlier, the model frequently regenerates it with a different identity (Fig.~\ref{fig:teaser}a).
Crucially, this behavior is not solely explained by limited context length; even when long histories are provided, models often treat distant observations as weak evidence.
In effect, they learn to \emph{continue} the immediate visual stream but fail to \emph{retrieve} decisive information from the distant past.

\begin{figure}[t]
    \centering
    \includegraphics[width=\textwidth]{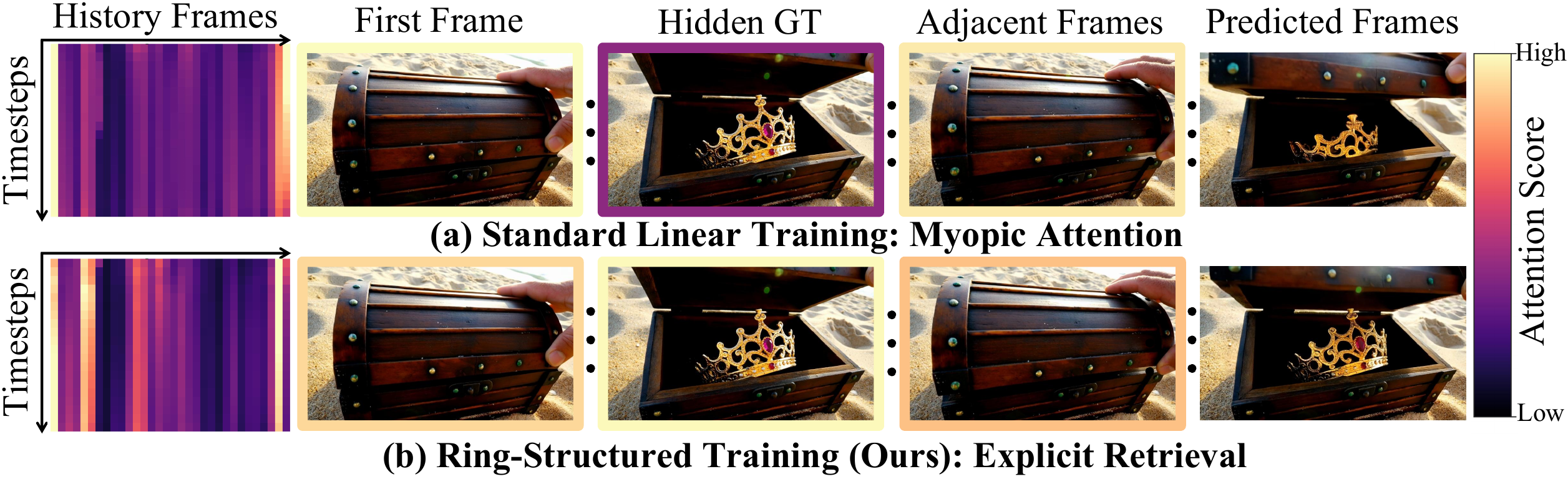}
    \caption{\textbf{Attention allocation of Predicted Frames to History Frames.} 
    \textbf{(a) Standard Linear Training} suffers from myopic attention bias, failing to retrieve effective information from distant history.
    \textbf{(b) Ring-Structured Training (Ours)} enables the model to extract more information from the past, achieving precise retrieval of historical frames.}
    \label{fig:mechanism}
    \vspace{-5mm}
\end{figure}

We identify the root cause of this limitation to a \textbf{\textit{misalignment between training objectives and inference requirements}}.
Standard video training data typically presents a ``linear narrative'' bias, where objects rarely exit and reappear within a short context window.
Constrained by this bias, models tend to learn ``myopic'' transition probabilities (i.e., inferring $x_{t+1}$ solely from $x_t$), with little incentive to learn how to retrieve information from distant history \textbf{(as visualized in Fig.~\ref{fig:mechanism}a)}.
In essence, the model learns to smoothly \textit{continue} the video sequence but fails to learn how to \textit{retrieve} answers from history, rendering long-term context effectively useless during inference even when provided.
Beyond the training paradigm, another fundamental bottleneck lies in the inability to precisely utilize distant historical information. 
Most existing long-video solutions operate at the frame or token level, creating an inherent dilemma: expanding the receptive field to cover long history requires increasing the sequence length, which incurs prohibitive computational costs; conversely, reducing the sequence length inevitably shrinks the receptive field.
This trade-off fundamentally limits the construction of long-term memory required for consistent long-video generation.

To address both bottlenecks, we propose \textbf{\name}, an autoregressive video diffusion framework designed for robust long-term memory construction and precise utilization.
Our core insight is to create training instances where the supervision for the current target is anchored in the distant past.
Specifically, we introduce a ring-structured data construction where ``the future becomes the history,'' embedding the ground-truth target clip into historical frames.
This turns long-range retrieval from an optional behavior into a necessity for minimizing training error, \textbf{yielding a highly precise retrieval focus on the distant target (Fig.~\ref{fig:mechanism}b)}.
To prevent degenerate shortcuts and to balance strict history adherence with open-ended continuation, we further employ a random head-cropping and context-drop strategy that controls whether the history contains direct supervisory evidence.

\textit{\name ~also targets scalability}. Leveraging the empirical prior that diffusion models emphasize global structure at high noise levels and fine details at low noise levels, we propose compression and timestep composition to expose complementary views of history across timesteps.
This design extends the effective history length to cover minutes-long durations under a fixed sequence-length budget while maintaining full historical coverage.
Finally, to maximize transfer from pretrained priors under variable compression and history length, we introduce sparse Rotary Positional Embedding (RoPE) that anchors compressed tokens to their physical spatiotemporal coordinates.
Experiments show that \name~ improves object permanence and enables faithful reuse of minutes-long history (Fig.~\ref{fig:teaser}), establishing a strong baseline for long-term memory in video generation.

Our main contributions are summarized as follows:
\vspace{-1mm}
\begin{itemize}
    \item We introduce {\name}, an autoregressive video diffusion framework that equips long-duration generation with precise long-term memory and improved object permanence.
    \item We devise ring-structured training that embeds targets into distant history and enforces explicit long-range retrieval, together with a cropping and context-drop mechanism that balances fidelity and diversity.
    \item We propose timestep-composed long-context conditioning integrated with sparse rotary positional encoding. This approach achieves a significant expansion in effective context under fixed sequence length constraints, enabling the faithful reproduction of minutes-long history and significantly outperforming SOTA methods. 
\end{itemize}

\section{Related Work}\label{sec:related}
\subsection{Autoregressive Video Diffusion}
While bidirectional video diffusion models such as Wan and HunyuanVideo~\cite{wan, hunyuan,hunyuan1.5,longcat} excel at short-clip generation, scaling them to long videos is constrained by the quadratic cost of bidirectional attention.
Consequently, autoregressive video diffusion has become the prevailing framework for long video generation. Recent works, including SkyReels-V2, Magi-1, CausVid and Self Forcing~\cite{skyreels, magi, causvid, selfforcing}, have achieved impressive visual quality. However, autoregressive generation inherently suffers from error accumulation and challenges in long-sequence context modeling, hindering the generation of longer and higher-quality videos.

To tackle this issue, various strategies have been proposed~\cite{diffusionforcing,hsg,st2v,far,streamdit}. One line of work focuses on training paradigms: FramePack~\cite{framepack} introduces a planned anti-drifting mechanism, while LongLive~\cite{longlive} employs attention sinks coupled with a ``train long, test long'' strategy. Other approaches enhance robustness by simulating inference errors during training. For instance, Rolling Forcing~\cite{rollingforcing} proposes a joint denoising scheme; Self-Forcing++~\cite{selfforcingpp} utilizes local teacher distillation on self-generated sequences; and SVI~\cite{svi} alongside Resampling Forcing~\cite{resampling} exposes the model to synthesized or imperfect history to learn error correction capabilities. In parallel, some works focus on efficient video generation architectures. For instance, SANA-Video~\cite{sana-video} incorporates linear attention into video generation, while others explore the use of more efficient VAEs~\cite{LTXV,DC-AE} or optimize attention mechanisms for long sequences~\cite{dao2022flashattention,dao2023flashattention2,zhang2025sageattention,zhang2024sageattention2,zhang2025sageattention3,radial-attn}. Furthermore, training-free methods have also been explored~\cite{tf1,tf2}, such as Deep Forcing~\cite{deepforcing}, which utilizes a ``Deep Sink'' mechanism, and Infinity-RoPE~\cite{infinityrope}, which adjusts Rotary Positional Embeddings to constrain the generation process closer to the pre-trained distribution.

\vspace{-2mm}
\subsection{Context Modeling for Long Video Generation}
Global consistency in long video generation relies on efficient context management, generally categorized into retrieval-based and compression-based methods.

\noindent \textbf{(1) Retrieval-based approaches} aim to select the most relevant historical information to extend the effective memory horizon. 
For instance, Context-as-Memory~\cite{contextasmemory} and WorldMem~\cite{worldmem} incorporate Field-of-View-based retrieval mechanisms within world models, while Pack-and-Force~\cite{packandforce} employs a contextual semantic retriever. 
Memory Forcing~\cite{memoryforcing} persists memory through 3D point cloud reconstruction, whereas Deep Forcing~\cite{deepforcing} utilizes attention mechanisms to retrieve relevant KV caches. RELIC~\cite{hong2025relic} uses time reversal to construct revisit data for
spatial memory in video world models.
Recent advancements also focus on hybrid architectures and state-space mechanisms; for example, VideoSSM~\cite{yu2025videossm} utilizes a hybrid State-Space Memory, and long-context state-space video world models~\cite{po2025long} have been developed to handle autoregressive long video generation efficiently. 
Furthermore, Mixture-of-Contexts~\cite{moc} learns attention routing to identify critical historical segments across multi-clip generation tasks. \textbf{(2) Compression-based methods} aim to condense historical information into compact representations. FramePack~\cite{framepack} compresses prior frames into fixed-size latent ``packs''.
WorldPack~\cite{oshima2025worldpack} improves spatial consistency in world models via history packing. TTTVideo~\cite{tttvideo} and LaCT~\cite{lact} introduce learnable parameters as memory representations that are updated during inference (Test-Time Optimization). Similarly, TinyHistory~\cite{pfp} pre-trains a context compression model via the reconstruction loss. 
Nevertheless, while these approaches successfully extend the input context window, they leave out direct training objectives that explicitly enforce the utilization and reproduction of historical information.

\vspace{-4mm}
\section{Method}\label{sec:method}
\begin{figure*}[t]
    \begin{center}
    \includegraphics[width=\textwidth]{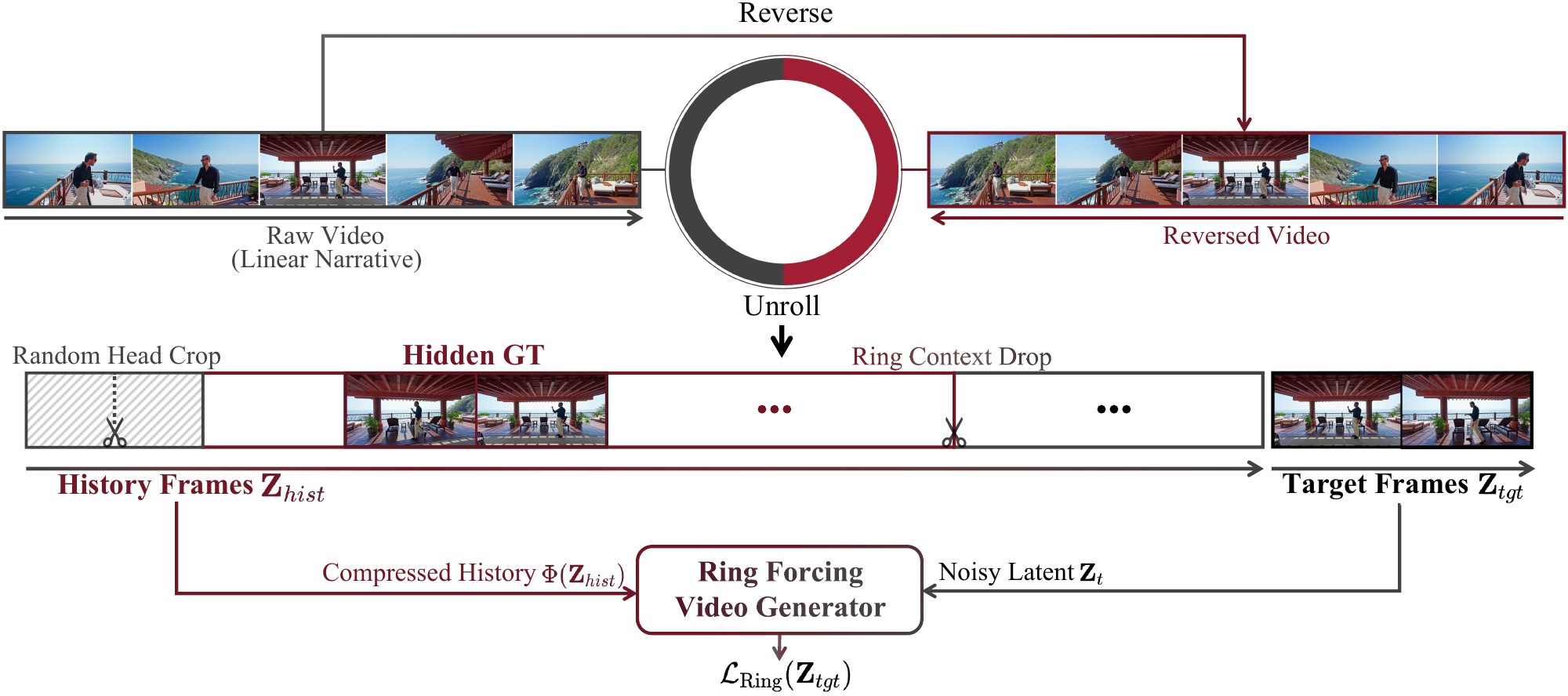}
    \end{center}
    \vspace{-0.2cm}
    \caption{\textbf{Overview of the Ring-Structured Training Strategy.} We construct a sequence ring by concatenating a linear raw video with its reversed counterpart. By unrolling the ring into a conditioning sequence, the target frames are naturally embedded into the distant history as the Hidden GT, which essentially forces the autoregressive generator to learn explicit long-range retrieval. To prevent trivial shortcuts, a Random Head Crop is applied to truncate boundary information leakage, while a Random Context Drop is used to balance history adherence and generation diversity.}
    \label{fig:ringforcing_strategy}
    \vspace{-5mm}
\end{figure*}

\vspace{-2mm}
We propose \name, an autoregressive video diffusion framework for long-term memory construction and precise utilization. It integrates three designs: ring-structured training for explicit long-range retrieval (Sec.~\ref{subsec:ring-data}), timestep-composed history compression under a fixed context budget (Sec.~\ref{subsec:timestep}), and sparse RoPE for consistent spatiotemporal scaling under variable compression (Sec.~\ref{subsec:sparseRoPE}). We train the model with a rectified-flow objective~\cite{esser2024scaling,liu2023flow} that unifies these components (Sec.~\ref{subsec:objective}).

\vspace{-4mm}
\subsection{Ring-Structured Training Strategy} 
\vspace{-1mm}
\label{subsec:ring-data}
As shown in Fig.~\ref{fig:ringforcing_strategy}, we introduce a training strategy based on \emph{ring-structured} sequences. The key idea is to embed the ground-truth target clip into the distant history, turning long-term retrieval into a self-supervised learning signal.
\vspace{-1mm}
\paragraph{Ring-Based Sequence Construction}
Formally, let $\mathcal{V} = \{v_1, v_2, \dots, v_N\}$ denote a raw video sequence of $N$ frames. We construct a closed-loop reference video $\mathcal{V}_{ring}$ by concatenating the original sequence with its reversed counterpart $\mathcal{V}_{rev} = \{v_N, v_{N-1}, \dots, v_1\}$:
\begin{equation}
    \mathcal{V}_{ring} = [\mathcal{V}, \mathcal{V}_{rev}].
\end{equation}

\paragraph{History-Target Decomposition}
We strictly sample the target video clip $\mathbf{x}_{tgt}$ from the forward part $\mathcal{V}$ to preserve the arrow of time. 
Based on the target sequence of length $L_{tgt}$, denoted as $\mathbf{x}_{tgt} = \{v_t, \dots, v_{t+L_{tgt}-1}\}$, we decompose $\mathcal{V}_{ring}$ into three consecutive segments:
(1) {Original history} $\mathbf{h}_{org} = \{v_1, \dots, v_{t-1}\}$,
(2) {Target clip} $\mathbf{x}_{tgt} = \{v_t, \dots, v_{t+L_{tgt}-1}\}$, and
(3) {Synthetic history} $\mathbf{h}_{syn}$ is the remaining segment in the ring, which contains the reversed target clip as a distant subsequence.

To construct the conditioning context, we place the synthetic history \emph{before} the original history:
\begin{equation}
    \mathbf{C}_{full} = [\mathbf{h}_{syn}, \mathbf{h}_{org}].
\end{equation}
This ring-based construction ensures visual continuity throughout the reference sequence by leveraging the temporal symmetry of the reversed video.

\paragraph{Leakage Prevention via Random Head Crop}
In this constructed history $\mathbf{C}_{full}$, a trivial shortcut (information leakage) exists at the head of the sequence. 
The first frame of $\mathbf{h}_{syn}$ is $v_{t+L_{tgt}}$, which is temporally adjacent and visually similar to the last frame of the target clip ($v_{t+L_{tgt}-1}$). 
If left unaddressed, the model can infer the end of the target simply by looking at the beginning of the history, ignoring the context (the reversed target clip $\mathbf{x}_{tgt}^{rev}$) embedded later in $\mathbf{h}_{syn}$. 
To prevent the leakage shortcut while forcing the model to leverage the full context, we apply a {Random Head Cropping} strategy.
Concretely, we randomly remove a prefix of the synthetic history while preserving the reversed target clip inside $\mathbf{h}_{syn}$, yielding the cropped context $\mathbf{C}_{crop}$, which removes the shortcut while keeping the distant answer retrievable.

\paragraph{Balancing Adherence and Diversity via Ring Context Drop}
To regulate the trade-off between strict history adherence (when the target is retrievable from the past) and open-ended generative diversity (when it is not), we introduce a \textbf{Ring Context Drop} mechanism. Rather than always enforcing the ring structure, we randomly drop the entire synthetic history during training. Specifically, we sample a boolean variable $b \sim \textit{Bernoulli}(1 - p_{\text{drop}})$ and formulate the final conditioning context as:
\begin{equation}
    \mathbf{C}_{final} = 
    \begin{cases} 
    \mathbf{C}_{crop}, & \text{if } b = 1 \quad (\text{Ring Mode}), \\
    \mathbf{h}_{org}, & \text{if } b = 0 \quad (\text{Standard Mode}).
    \end{cases}
\end{equation}
When $b=0$, the ring-constructed context is completely dropped, thereby preventing the model from over-relying on the synthetic history and forcing it to learn unconstrained progression solely from $\mathbf{h}_{org}$. When $b=1$, the model learns to explicitly leverage the embedded target for precise long-term consistency.

\vspace{-2mm}
\subsection{Compression and Timestep Composition} \label{subsec:timestep}
\vspace{-1mm}
While the Ring-Structured strategy provides the essential supervision for long-range retrieval, directly attending to minute-long raw history is computationally prohibitive due to the quadratic complexity of attention. 
To resolve this, we propose \textbf{Timestep-Composed Compression}, which maps a long history $\mathbf{Z}_{hist}$ into a bounded conditioning sequence $\tilde{\mathbf{C}}$ under a fixed token budget $\mathcal{B}_{max}$. This design ensures efficient memory utilization while preserving both global structure and local details.
\vspace{-3mm}
\paragraph{Token Budget and Compression Operator}
We define a spatiotemporal compression operator $\Psi(\mathbf{Z}; r_h, r_w, r_t)$ that downsamples history in space and time to fit $\mathcal{B}_{max}$. 
\vspace{-3mm}

\paragraph{Timestep-Dependent Composition Operator}
Uniform spatiotemporal downsampling inevitably sacrifices high-frequency details. However, we leverage the intrinsic generative nature of diffusion models: \emph{high-noise intervals primarily determine semantic structure and global motion, whereas low-noise intervals focus on refining local textures and edges.}
Motivated by this, we define a \textbf{Composite Condition Operator} $\Phi(\mathbf{Z}_{hist}, t, \delta)$ that dynamically transforms the history representation based on the diffusion timestep $t$:

\begin{equation}
    \Phi(\mathbf{Z}_{hist}, t, \delta) = 
    \begin{cases} 
        \Psi(\mathbf{Z}_{hist}; r_h^{\text{g}}, r_w^{\text{g}}, r_t^{\text{g}}), & t > \tau \\
        \mathcal{S}\left(\Psi(\mathbf{Z}_{hist}; r_h^{\text{d}}, r_w^{\text{d}}, r_t^{\text{d}}), \delta\right), & t \le \tau
    \end{cases}
\end{equation}

Here, the first branch provides a temporally dense context for recovering global structure, while the second branch preserves high-frequency spatial details with temporally sparse sampling. $\mathcal{S}(\cdot, \delta)$ is a cyclic shift over the temporal sampling grid, and varying $\delta$ ensures full coverage over long histories.

\Needspace{18\baselineskip}
\begin{wrapfigure}[14]{l}{0.50\textwidth}
\vspace{-9mm}
\centering
\includegraphics[width=\linewidth]{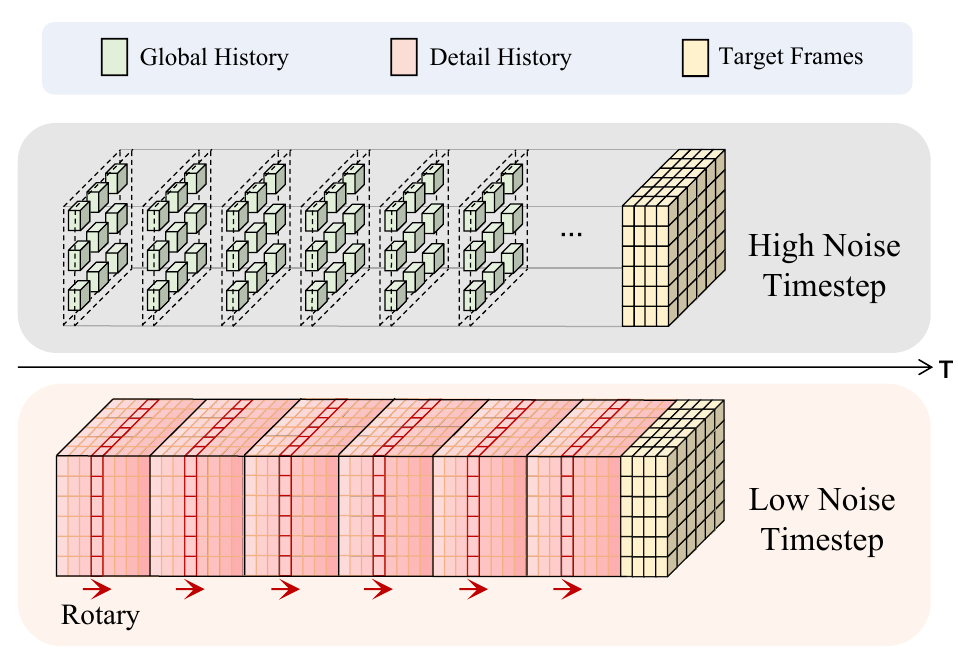}
\vspace{-3mm}
\caption{\textbf{Timestep-composed long-context conditioning.}}
\label{fig:sparse_com}
\vspace{-4mm}
\end{wrapfigure}

\subsection{Sparse RoPE Strategy}  \label{subsec:sparseRoPE}
To maximize the use of pretrained video generation priors (i.e., the model's perception of relative space-time structures), and to accommodate historical contexts of varying compression rates and lengths, we introduce a sparse RoPE strategy.
As shown in Fig.~\ref{fig:sparse_com}, instead of encoding the compressed history features using their discrete indices in the latent space, we map them back to the original continuous spatiotemporal coordinate system used during pre-training.

Formally, we treat the target clip as the reference coordinate system with origin $(0,0,0)$. For a compressed history token at index $(t,h,w)$ with spatial downsampling factor $s_s$ and temporal downsampling factor $s_t$, we map it back to physical coordinates $(\tilde{t},\tilde{h},\tilde{w})$ in the original grid:
\begin{align}
\tilde{h} &= h\cdot s_s + 0.5\cdot (s_s-1),\nonumber\\
\tilde{w} &= w\cdot s_s + 0.5\cdot (s_s-1),\\
\tilde{t} &= -\left((L_{hist}-1-t)\cdot s_t + 0.5\cdot (s_t-1) + 0.5\right),
\end{align}
where $L_{hist}$ is the number of compressed history steps along time. The spatial terms center the token within its corresponding patch, and the temporal term places the history strictly in the negative relative-time domain. We then compute 3D RoPE using $(\tilde{t},\tilde{h},\tilde{w})$ to preserve consistent relative spatiotemporal scales under variable compression.

\subsection{Training Objective} \label{subsec:objective}
We adopt the rectified flow framework and train the generator $G_\theta$ to predict the velocity field derived from the clean target $\mathbf{Z}_{tgt}$. The training objective unifies the ring-structured strategy (controlled by $b$) and the compression strategy (encapsulated by $\Phi$) into a holistic loss function.
Let $\mathbf{Z}_{hist}^{(b)}$ denote the history latents constructed by the ring strategy, where $b \sim \text{Bernoulli}(1-p_{\text{drop}})$ determines whether to use the ring context ($b=1$) or standard context ($b=0$). The final conditioning input is computed as $\tilde{\mathbf{C}} = \Phi(\mathbf{Z}_{hist}^{(b)}, t, \delta)$.
The formal optimization objective is:
\begin{equation}
    \mathcal{L}_{\text{Ring}} = \mathbb{E}_{t, \mathbf{Z}_{tgt}, \epsilon, b, \delta} \left[ \left\| (\epsilon - \mathbf{Z}_{tgt}) - G_\theta(\mathbf{Z}_t, t, \tilde{\mathbf{C}}) \right\|^2_2 \right].
\end{equation}

In this formulation, $t \sim \mathcal{U}(0,1)$ represents the sampling timestep, and $\mathbf{Z}_t = (1-t)\mathbf{Z}_{tgt} + t\epsilon$ denotes the noisy latent mixed with Gaussian noise $\epsilon \sim \mathcal{N}(\mathbf{0}, \mathbf{I})$. The joint expectations over $b$, $t$, and $\delta$ optimize the model to balance precise history retrieval with generative diversity, while efficiently capturing both global structures and local details across varying noise levels.


\section{Experiments}\label{sec:exp}
In this section, we present a comprehensive evaluation of \name{}. We begin by detailing the implementation and experimental setup. Next, we compare \name{} against state-of-the-art long video generation models to demonstrate its superiority in maintaining object permanence and long-term consistency. Finally, we conduct in-depth ablation studies to validate the efficacy of our core contributions: the Timestep Composition strategy, the ring-structured training trade-offs, and the sparse RoPE mechanism.

\subsection{Implementation Details}
\paragraph{Dataset Construction}
We utilize the UltraVideo-Long dataset~\cite{xue2025ultravideo}. 
To isolate the challenge of temporal consistency within continuous shots, we filter the dataset to retain only single-shot videos. We curate a training set of {10,000} long-duration videos, each with a duration ranging from 10 to 240 seconds. All video data is standardized to a resolution of $480 \times 832$ and a frame rate of 15 FPS to match the training requirements of the base models.
\paragraph{Model Instantiation} We instantiate \name{} using two backbones: Wan2.1 T2V 1.3B and Wan2.2 T2V A14B~\cite{wan}. All ablation studies are conducted on the Wan2.1 1.3B variant, while the final comparative evaluation employs the Wan2.2 A14B model. To ensure parameter efficiency, all models are fine-tuned using Low-Rank Adaptation (LoRA)~\cite{hu2022lora} with a rank of $r=128$. 
Training for the Wan2.2 A14B model was distributed across a cluster of 32 NVIDIA H800 GPUs for 7,000 steps, taking approximately 30 hours. For inference, we utilize a single NVIDIA H800 GPU. By aligning the effective sequence length with the standard generation length via compression, our method maintains computational costs virtually identical to the original model. Consequently, \textit{\name{} is deployable on any hardware supporting the base Wan model, incurring negligible overhead in VRAM or compute.}
\vspace{-2mm}


\subsection{Comparison with Baselines}
\paragraph{Baselines} We compare \name{} against three leading long-video generation models: LongLive~\cite{longlive}, LongCat~\cite{longcat}, and FramePack~\cite{framepack}.

\paragraph{Evaluation Protocol}
Object permanence is a core manifestation of a video generation model's long-term memory capacity. To rigorously quantify this capability, we specifically design a benchmark based on an ``Appear-Disappear-Reappear'' (A-D-R) logic. Additionally, to evaluate the model's long video generation capabilities, we conduct a targeted assessment of its general performance in generating 1-minute coherent long videos.
\begin{table*}[t]
\centering
\caption{\textbf{Quantitative results on the A-D-R Benchmark.} Metrics are categorized into Consistency and General Performance. \textbf{I.F.} denotes Instruction Following. For $\Delta t_{gap} > 5s$, baselines are excluded because the ``appear'' segment falls completely outside their maximum context window. Best results are highlighted in bold, and second-best are underlined.}
\vspace{-1mm}
\label{tab:adr_results}

\footnotesize
\setlength{\tabcolsep}{3pt}
\renewcommand{\arraystretch}{1.15}

\begin{adjustbox}{width=\textwidth,center}
\begin{tabularx}{\textwidth}{c l *{6}{>{\centering\arraybackslash}X}}
\toprule
\multirow{2}{*}{\makecell[c]{\textbf{Gap}\\($\Delta t_{gap}$)}} &
\multirow{2}{*}{\textbf{Model}} &
\multicolumn{3}{c}{\textbf{Consistency}} &
\multicolumn{3}{c}{\textbf{General Performance}} \\
\cmidrule(lr){3-5}\cmidrule(lr){6-8}
& & Texture & Geometry & Semantic & Aesthetic & I.F. & Dynamics \\
\midrule

\multirow{4}{*}{\textbf{0s}}
& LongLive  & 0.578 & 0.896 & 0.886 & 5.032 & \underline{24.117} & \underline{1.615} \\
& LongCat   & \underline{0.737} & \underline{0.950} & \textbf{0.925} & 5.214 & 24.090 & 1.493 \\
& FramePack & 0.605 & 0.949 & 0.913 & \textbf{5.222} & 23.162 & \textbf{1.662} \\
& \textbf{Ours} & \textbf{0.742} & \textbf{0.997} & \underline{0.918} & \underline{5.216} & \textbf{24.642} & 1.550 \\
\midrule

\multirow{4}{*}{\textbf{1s}}
& LongLive  & 0.181 & 0.297 & 0.707 & 5.080 & 23.803 & 1.694 \\
& LongCat   & 0.236 & \underline{0.368} & \underline{0.712} & 5.138 & \underline{23.894} & \underline{1.990} \\
& FramePack & \underline{0.252} & 0.287 & 0.692 & \textbf{5.220} & 23.040 & 1.093 \\
& \textbf{Ours} & \textbf{0.723} & \textbf{0.839} & \textbf{0.823} & \underline{5.176} & \textbf{24.601} & \textbf{2.617} \\
\midrule

\multirow{4}{*}{\textbf{5s}}
& LongLive  & \underline{0.290} & 0.303 & \underline{0.730} & 5.065 & \underline{23.933} & 1.555 \\
& LongCat   & 0.203 & \underline{0.328} & 0.722 & 5.148 & 23.859 & \underline{1.640} \\
& FramePack & 0.250 & 0.264 & 0.701 & \textbf{5.229} & 22.795 & 0.996 \\
& \textbf{Ours} & \textbf{0.515} & \textbf{0.761} & \textbf{0.831} & \underline{5.163} & \textbf{23.940} & \textbf{2.371} \\
\midrule

\rowcolor{mygray}\textbf{15s} & \textbf{Ours} & 0.509 & 0.653 & 0.802 & 5.104 & 23.803 & 2.689 \\
\rowcolor{mygray}\textbf{30s} & \textbf{Ours} & 0.421 & 0.621 & 0.822 & 5.055 & 23.590 & 2.946 \\
\rowcolor{mygray}\textbf{60s} & \textbf{Ours} & 0.396 & 0.582 & 0.793 & 5.014 & 22.872 & 1.809 \\
\bottomrule
\end{tabularx}
\end{adjustbox}

\end{table*}
\paragraph{A-D-R Benchmark}
To rigorously and fairly measure the model's memory retrieval capability across temporal dimensions, we meticulously construct an A-D-R prompt library comprising 64 test cases. Each test case consists of four core elements: an appear prompt, a disappear prompt, a reappear prompt, and a precise subject keyword. During video generation, the durations of the ``appear'' and ``reappear'' segments are fixed at 5 seconds each. Furthermore, to prevent the model from exploiting shortcuts by relying solely on the initial frame, the first frame is intentionally devoid of the target subject, and its appearance is randomly triggered within the first 5 seconds, thereby better approximating realistic dynamic generation scenarios. By dynamically adjusting the duration of the intermediate ``disappear'' segment, we systematically investigate the model's capability bounds in maintaining subject identity across varying temporal spans. Specifically, we set the disappearance durations to 0 (where the entire video is evaluated as a control group), 1, 5, 15, 30 and 60 seconds.

To maximally isolate the consistency evaluation from background interference, we employ SAM~3~\cite{carion2025sam} combined with subject keywords to perform precise zero-shot instance segmentation on the appear and reappear segments. We also perform subject detection on the disappear segment to exclude cases where the subject fails to vanish from consistency calculations, while our meticulous prompt design ensures that most cases successfully adhere to the A-D-R logic. For both segments, we extract the keyframe with the largest subject mask area, crop the subject, and normalize it against a pure white background. For the extracted subject images before and after disappearance, we comprehensively employ SIFT~\cite{lowe2004SIFT}, LoFTR~\cite{sun2021loftr}, and DINOv3~\cite{simeoni2025dinov3} to evaluate local texture consistency, geometric structure consistency, and high-level semantic consistency, respectively. This multi-dimensional feature matching strategy allows us to thoroughly quantify the preservation of object identity. Furthermore, general metrics are evaluated on the reappear segments: we utilize the Improved Aesthetic Predictor~\cite{schuhmann2022improved} to quantify the aesthetic quality of the videos, employ X-CLIP~\cite{XCLIP} to measure instruction following capabilities, and apply Optical Flow-based Motion Magnitude to quantify the amplitude of video dynamics.

\begin{figure*}[t]
    \begin{center}
    \includegraphics[width=1.0\textwidth]{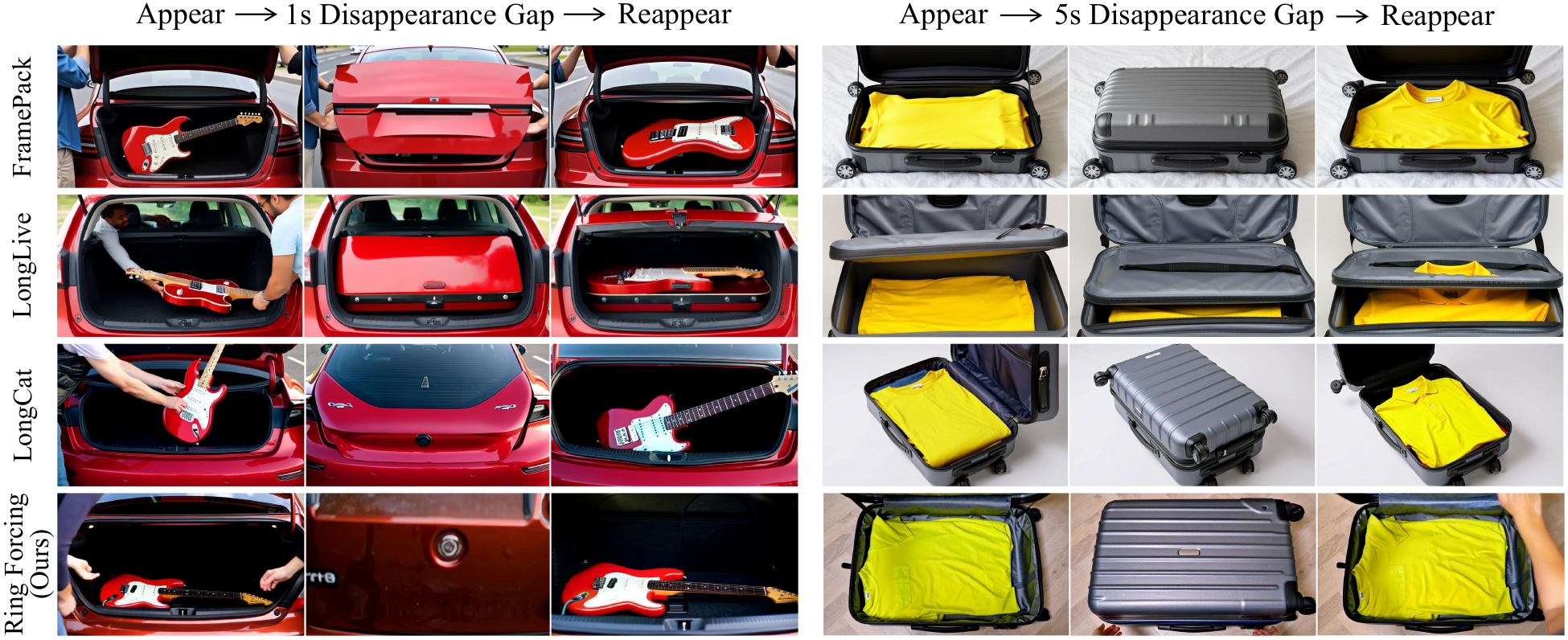}
    \end{center}
    \vspace{-0.2cm}
    \caption{Qualitative results on the Appear–Disappear–Reappear (A-D-R) benchmark. Our model demonstrates robust object permanence by retrieving identity information from distant history, ensuring consistent subject reconstruction even after long temporal gaps where baseline models typically fail.}
    \label{fig:adr}
    \vspace{-3mm}
\end{figure*}
\paragraph{General Video Generation Benchmark}
To comprehensively evaluate the model's generalized capability in regular scenarios, we construct a General Benchmark comprising 64 diverse prompts. These prompts are utilized to generate 60-second continuous videos where the explicit ``disappear-reappear'' logic is absent, and the primary subjects maintain a persistent presence. We adopt the same automated metrics used in the A-D-R benchmark to evaluate aesthetic quality, prompt adherence, and motion magnitude, and further include \textit{Motion Smoothness} by calculating the normalized frame interpolation error via AMT~\cite{AMT} to quantify temporal fluidity. Furthermore, to assess complex perceptual qualities that automated metrics struggle to capture, we introduce the Large Multimodal Model Qwen3-VL~\cite{Qwen3-VL} to conduct a comprehensive 5-point scale evaluation. This evaluation focuses on three critical dimensions: \textit{physical logic rationality} to ensure realistic object interactions, \textit{spatiotemporal consistency} to prevent semantic collapse over time, and \textit{visual naturalness} to penalize unnatural flickering and morphological distortions.

\paragraph{Human Evaluation}
To further validate our results through human perception, we conducted a Human Evaluation involving 21 participants from diverse backgrounds. We randomly sampled 10 video pairs from the General Benchmark (pairing our generated videos with those from baseline models). Under a strict blind-test protocol, evaluators rated each video on a 0-5 scale with a primary focus on ``Overall Video Quality.''

\begin{table*}[t]
    \centering
    \caption{\textbf{Quantitative comparison on the General long-video generation benchmark (1 minute).}}
    \label{tab:general_comparison}
    \small
    \setlength{\tabcolsep}{4pt}
    \renewcommand{\arraystretch}{1.15}
    \resizebox{0.95\textwidth}{!}{
        \begin{tabular}{l ccccc cc}
            \toprule
            \multirow{2}{*}{\textbf{Model}} 
            & \multicolumn{5}{c}{\textbf{Objective Metrics}} 
            & \multicolumn{2}{c}{\textbf{User Study}} \\
            \cmidrule(lr){2-6} \cmidrule(lr){7-8}
            & \makecell[c]{Aesthetic} 
            & \makecell[c]{Dynamics} 
            & \makecell[c]{Naturalness} 
            & \makecell[c]{Motion\\Smoothness} 
            & \makecell[c]{Instruction\\Following} 
            & \makecell[c]{Consistency} 
            & \makecell[c]{Overall\\Quality} \\
            \midrule
            LongLive  
            & 5.622 
            & 1.056 
            & 3.031 
            & \textbf{0.991} 
            & 24.805 
            & \underline{4.12} 
            & 2.25 \\

            LongCat   
            & \underline{5.808} 
            & \underline{2.196} 
            & \underline{3.875} 
            & \textbf{0.991} 
            & \underline{25.341} 
            & 3.59 
            & \underline{4.19} \\

            FramePack 
            & 5.728 
            & 1.790 
            & 3.656 
            & \textbf{0.991} 
            & 25.014 
            & 3.71 
            & 4.04 \\

            \rowcolor{mygray} \textbf{Ours} 
            & \textbf{5.822} 
            & \textbf{2.261} 
            & \textbf{3.922} 
            & \underline{0.986} 
            & \textbf{25.471} 
            & \textbf{4.35} 
            & \textbf{4.22} \\
            \midrule

            Base model$^\dagger$ 
            & 5.825 
            & 1.974 
            & 4.016 
            & 0.984 
            & 25.136 
            & \multicolumn{2}{c}{\textit{N/A}} \\
            \bottomrule
        \end{tabular}
    }
    \vspace{-2pt}
    \footnotesize\textit{$^\dagger$Wan2.2-T2V-A14B (5.4s).}
    \vspace{-4mm}
\end{table*}

\begin{figure*}[h]
    \begin{center}
    \includegraphics[width=\textwidth]{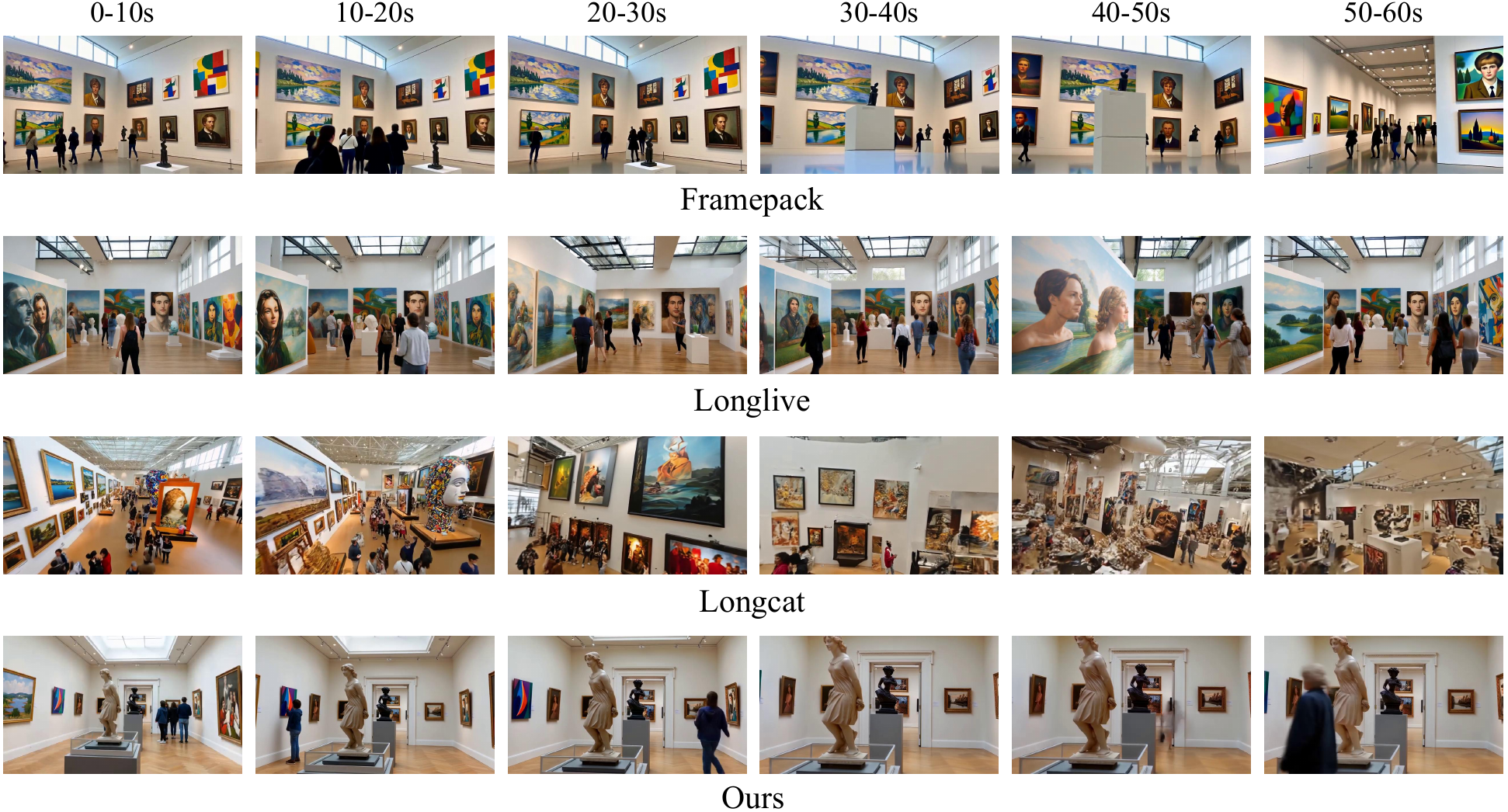}
    \end{center}
    \vspace{-0.2cm}
\caption{Qualitative results of 60-second long video generation. The visualization demonstrates the model's capability to maintain overall consistency and visual quality throughout a 1-minute duration in general scenarios.}
\label{fig:com}
\end{figure*}

\paragraph{Results Analysis}
Comprehensive evaluations demonstrate that \name{} achieves state-of-the-art long-term memory and general video generation quality. 
First, on the A-D-R benchmark (Tab.~\ref{tab:adr_results}, Fig.~\ref{fig:adr}), \name{} exhibits robust object permanence by precisely retrieving distant history. Conversely, baselines suffer catastrophic forgetting when targets temporarily disappear, restricted by short-sighted attention biases from linear training.
Second, \name{} fundamentally breaks the ``consistency vs. dynamics'' trade-off. Baselines like LongLive over-rely on the initial frame as an attention sink, degenerating into static or repetitive outputs (low Dynamics). In contrast, \name{} yields rich, fluid motions while maintaining strict spatiotemporal consistency.
Finally, General Benchmark results (Tab.~\ref{tab:general_comparison}, Fig.~\ref{fig:com}) show that \name{} preserves the backbone's generative priors while extending generation to 60-second videos. Among long-video baselines, it achieves the best aesthetics, naturalness, and overall human preference, demonstrating that our long-term memory mechanism improves temporal coherence without substantially degrading visual quality.

\subsection{Ablation Studies}
\paragraph{Impact of Compression and Timestep Composition}
We first investigate the impact of spatiotemporal compression strategies on the model's ability to reconstruct history. We construct a test set of {128 unseen samples} where the ``answer'' (ground truth target) is naturally embedded within the history using our Ring-Structured construction. We evaluate reconstruction fidelity using photometric and perceptual metrics~\cite{zhang2018unreasonable}.

\begin{wrapfigure}{l}{6.4cm}
\vspace{-.2cm}
\centering
\includegraphics[width=0.5\textwidth]{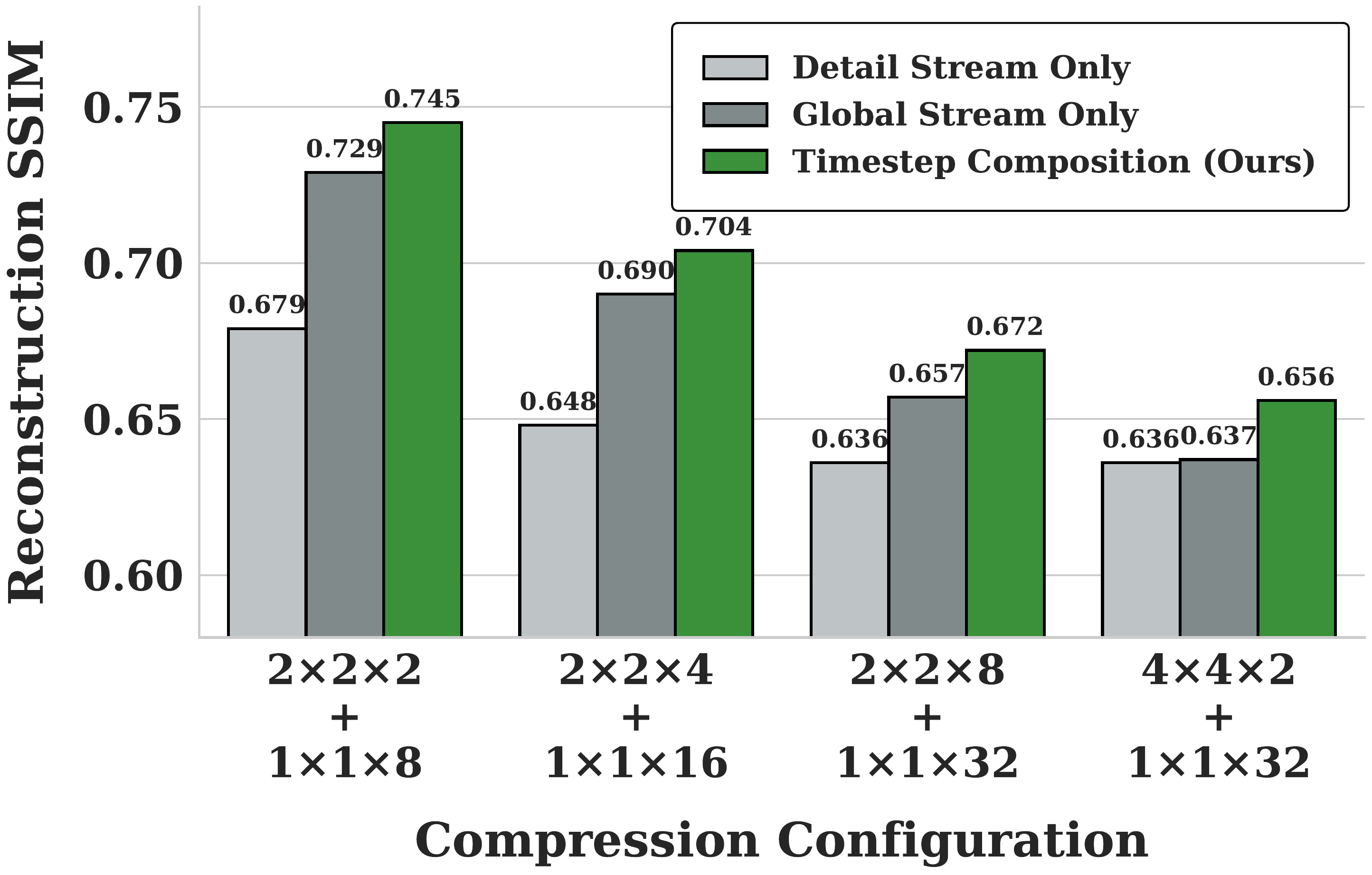}
\vspace{-0.0cm}
\caption{\textbf{Effectiveness of Timestep Composition Strategy.}}
\vspace{-0.4cm}
\label{fig:ablation_composition}
\end{wrapfigure} 

\paragraph{Compression Trade-offs and Composition Effectiveness}
As illustrated in Fig. ~\ref{fig:ablation_composition}, we first analyze the single-stream baselines (gray bars) to understand the trade-offs between spatial and temporal information. We observe that the ``Global Stream Only'' baseline (high spatial compression, intact temporal density) generally yields superior reconstruction SSIM compared to the ``Detail Stream Only'' baseline (aggressive temporal compression). This suggests that for precisely leveraging long-term history, retaining temporal density is often more critical than spatial resolution.
Building on this insight, we evaluate our proposed timestep composition strategy. As shown by the green bars in Fig. ~\ref{fig:ablation_composition}, our method dynamically synergizes the structural guidance from the global stream with the fine-grained refinement from the detail stream. Remarkably, without increasing the sequence length compared to single-stream baselines, the combined strategy consistently outperforms both standalone baselines across all metrics and compression settings, compellingly validating the effectiveness of our dual-stream design.

To balance reconstruction quality and long-video capacity, we select $2\times2\times8 + 1\times1\times32$ as our default. Under this configuration, the inference overhead for 140s of history is virtually identical to the original Wan model’s 5.4s generation, enabling efficient long-video synthesis without increasing hardware requirements.
\vspace{-3mm}
\paragraph{Trade-offs of Adherence and Diversity}
To determine the optimal balance between object permanence and generative diversity, we conduct an ablation on the Ring Context Drop probability $p_{drop}$ (which controls the Bernoulli variable $b$). We employ a dual-evaluation protocol that measures reconstruction fidelity on 128 unseen samples with embedded answers (Ring Mode), and generates 15-second video continuations on 128 samples without embedded answers (Standard Mode). In this mode, we employ Qwen3-VL to assess the diversity of the generated content relative to the input history on a 5-point Likert scale, and we report the average score across all samples. As shown in Tab.~\ref{tab:ablation_pdrop}, we identify $p_{drop}=0.4$ as the ``sweet spot'', where the model maintains high reconstruction fidelity without collapsing into mode dropping.

\begin{table}[t]
    \centering
    \small
    \caption{Ablation study on Ring Context Drop probability $p_{drop}$. }
    \label{tab:ablation_pdrop}
 
    \begin{tabular}{c ccc c}
    \toprule
    \multirow{2}{*}{\textbf{$p_{drop}$}} & \multicolumn{3}{c}{\textbf{Ring Mode (Reconstruction)}} & \multicolumn{1}{c}{\textbf{Standard Mode}} \\
    \cmidrule(lr){2-4} \cmidrule(lr){5-5}
     & \textbf{PSNR} $\uparrow$ & \textbf{SSIM} $\uparrow$ & \textbf{LPIPS} $\downarrow$ & \textbf{Diversity Score} $\uparrow$ \\
    \midrule
    0.0 & \textbf{25.02} & \textbf{0.73} & \textbf{0.12} & 3.71 \\
    0.2 & \underline{24.67} & \underline{0.72} & \textbf{0.12} & 3.93 \\
    0.4 & 24.22 & 0.71 & \underline{0.13} & 4.11 \\
    0.6 & 24.02 & 0.71 & \underline{0.13} & 4.13 \\
    0.8 & 23.02 & 0.68 & 0.16 & \underline{4.21} \\
    1.0 & 21.86 & 0.65 & 0.21 & \textbf{4.23} \\
    \bottomrule
    \end{tabular}
\end{table}
\vspace{-2mm}
\paragraph{Efficacy of Sparse RoPE} To validate whether our Sparse RoPE strategy effectively leverages pre-trained priors, we compare the reconstruction capabilities of two models trained with identical settings: one using standard index-based RoPE and the other using our Sparse RoPE. As presented in Tab.~\ref{tab:sparse_rope}, experimental results demonstrate that sparse RoPE better utilizes priors to achieve faster convergence and superior reconstruction quality. This confirms that mapping compressed tokens back to their physical coordinates significantly aids the model in understanding relative spatiotemporal relationships.
\begin{table}[t]
    \centering
    \small
    \caption{Comparison of reconstruction capabilities between standard index-based RoPE and our sparse RoPE. }
    \label{tab:sparse_rope}
    \begin{tabular}{l ccc}
    \toprule
    \textbf{Method} & \textbf{PSNR $\uparrow$} & \textbf{SSIM $\uparrow$} & \textbf{LPIPS $\downarrow$} \\
    \midrule
    Standard RoPE & 19.05 & 0.58 & 0.27 \\
    \rowcolor{mygray} \textbf{Sparse RoPE (Ours)} & \textbf{24.22} & \textbf{0.71} & \textbf{0.13} \\
    \bottomrule
    \end{tabular}
    \vspace{-3mm}
\end{table}

\vspace{-2mm}
\section{Conclusion}\label{sec:conclu}
\vspace{-2mm}
We address the challenge of establishing robust long-term memory in autoregressive video generation, specifically targeting the critical bottlenecks of object permanence and memory capacity.
We proposed \name{}, a unified framework that enables the robust construction and precise utilization of long-term memory.
Through our ring-structured training strategy, we successfully extracted rich information from long-term history, effectively reconciling the trade-off between historical adherence and generative diversity.
Furthermore, our compression and timestep composition mechanism expanded the effective historical span to support minutes-long generation under fixed sequence length constraints, achieving a comprehensive receptive field over the entire history. Together with the sparse RoPE mechanism, our approach enables the precise utilization and reproduction of information spanning minutes-long history. \name{} establishes a foundational paradigm for future research into infinite-context generative modeling, paving the way for consistent, long-duration video synthesis.

\bibliographystyle{splncs04}
\bibliography{main}

@String(CVPR  = {Conference on Computer Vision and Pattern Recognition})

@String(ICCV  = {International Conference on Computer Vision})

@String(NeurIPS  = {Conference and Workshop on Neural Information Processing Systems})

@String(ICLR  = {International Conference on Learning Representations})

@String(ICML = {International Conference on Machine Learning})

@String(ECCV  = {European Conference on Computer Vision})

@String(ACMMM = {{ACM} International Conference on Multimedia})

@misc{sora2,
  title        = {Sora 2 is here},
  author       = {OpenAI},
  year         = {2025},
  url          = {https://openai.com/index/sora-2/},
  note         = {Accessed: 2026-06-30},
}

@misc{gemini_video_generation,
  title        = {Veo 3.1},
  author       = {{Google DeepMind}},
  year         = {2025},
  url          = {https://deepmind.google/models/veo/},
  note         = {Accessed: 2026-06-30},
}

@article{kling_ai_global,
      title={Kling-Omni Technical Report}, 
      author={{Kling Team}},
      year={2025},
  journal = {ArXiv preprint},
}

@article{wan,
  author       = {{Wan Team}},
  title        = {Wan: Open and Advanced Large-Scale Video Generative Models},
  journal = {ArXiv preprint},
  year         = {2025},
}

@article{hunyuan,
  author       = {{Hunyuan Foundation Model Team}},
  title        = {HunyuanVideo: {A} Systematic Framework For Large Video Generative
                  Models},
  journal      = {ArXiv preprint},
  year         = {2024},
}

@article{hunyuan1.5,
  author       = {{Hunyuan Foundation Model Team}},
  title        = {HunyuanVideo 1.5 Technical Report},
  journal      = {ArXiv preprint},
  year         = {2025},
}

@article{skyreels,
  author       = {{SkyReels Team}},
  title        = {SkyReels-V2: Infinite-length Film Generative Model},
  year         = {2025},
  journal      = {ArXiv preprint},

}

@article{magi,
  author       = {{Sand AI}},
  title        = {{MAGI-1:} Autoregressive Video Generation at Scale},
  journal      = {ArXiv preprint},
  year         = {2025},
}

@inproceedings{causvid,
  author       = {Tianwei Yin and
                  Qiang Zhang and
                  Richard Zhang and
                  William T. Freeman and
                  Fr{\'{e}}do Durand and
                  Eli Shechtman and
                  Xun Huang},
  title        = {From Slow Bidirectional to Fast Autoregressive Video Diffusion Models},
  booktitle    = CVPR,
  year         = {2025},
}

@inproceedings{selfforcing,
  title={Self Forcing: Bridging the Train-Test Gap in Autoregressive Video Diffusion},
  author={Huang, Xun and Li, Zhengqi and He, Guande and Zhou, Mingyuan and Shechtman, Eli},
 booktitle=NeurIPS,
  year={2025}
}

@inproceedings{framepack,
    title={Frame Context Packing and Drift Prevention in Next-Frame-Prediction Video Diffusion Models},
    author={Lvmin Zhang and Shengqu Cai and Muyang Li and Gordon Wetzstein and Maneesh Agrawala},
    booktitle=NeurIPS,
    year={2025},
}

@article{longlive,
  author       = {Shuai Yang and
                  Wei Huang and
                  Ruihang Chu and
                  Yicheng Xiao and
                  Yuyang Zhao and
                  Xianbang Wang and
                  Muyang Li and
                  Enze Xie and
                  Yingcong Chen and
                  Yao Lu and
                  Song Han and
                  Yukang Chen},
  title        = {LongLive: Real-time Interactive Long Video Generation},
  journal      = {ArXiv preprint},
  year         = {2025},
}

@article{rollingforcing,
  author       = {Kunhao Liu and
                  Wenbo Hu and
                  Jiale Xu and
                  Ying Shan and
                  Shijian Lu},
  title        = {Rolling Forcing: Autoregressive Long Video Diffusion in Real Time},
  journal      = {ArXiv preprint},
  year         = {2025},
}

@article{selfforcingpp,
  author       = {Justin Cui and
                  Jie Wu and
                  Ming Li and
                  Tao Yang and
                  Xiaojie Li and
                  Rui Wang and
                  Andrew Bai and
                  Yuanhao Ban and
                  Cho{-}Jui Hsieh},
  title        = {Self-Forcing++: Towards Minute-Scale High-Quality Video Generation},
  journal      = {ArXiv preprint},
  year         = {2025},
}

@article{svi,
  author       = {Wuyang Li and
                  Wentao Pan and
                  Po{-}Chien Luan and
                  Yang Gao and
                  Alexandre Alahi},
  title        = {Stable Video Infinity: Infinite-Length Video Generation with Error
                  Recycling},
  journal      = {ArXiv preprint},
  year         = {2025},
}

@article{resampling,
    title={End-to-End Training for Autoregressive Video Diffusion via Self-Resampling}, 
    author={Yuwei Guo and Ceyuan Yang and Hao He and Yang Zhao and Meng Wei and Zhenheng Yang and Weilin Huang and Dahua Lin},
    year={2025},
  journal      = {ArXiv preprint},
}

@article{deepforcing,
    title={Deep Forcing: Training-Free Long Video Generation with Deep Sink and Participative Compression},
    author={Yi, Jung and Jang, Wooseok and Cho, Paul Hyunbin and Nam, Jisu and Yoon, Heeji and Kim, Seungryong},
  journal      = {ArXiv preprint},
    year={2025}
}

@article{infinityrope,
      title={Infinity-RoPE: Action-Controllable Infinite Video Generation Emerges From Autoregressive Self-Rollout}, 
      author={Hidir Yesiltepe and Tuna Han Salih Meral and Adil Kaan Akan and Kaan Oktay and Pinar Yanardag},
      year={2025},
  journal      = {ArXiv preprint},
}

@article{contextasmemory,
  author       = {Jiwen Yu and
                  Jianhong Bai and
                  Yiran Qin and
                  Quande Liu and
                  Xintao Wang and
                  Pengfei Wan and
                  Di Zhang and
                  Xihui Liu},
  title        = {Context as Memory: Scene-Consistent Interactive Long Video Generation
                  with Memory Retrieval},
  journal      = {ArXiv preprint},
  year         = {2025},
}

@inproceedings{worldmem,
  author       = {Zeqi Xiao and
                  Yushi Lan and
                  Yifan Zhou and
                  Wenqi Ouyang and
                  Shuai Yang and
                  Yanhong Zeng and
                  Xingang Pan},
  title        = {{WORLDMEM:} Long-term Consistent World Simulation with Memory},
  year         = {2025},
    booktitle=NeurIPS,
}

@article{packandforce,
  author       = {Xiaofei Wu and
                  Guozhen Zhang and
                  Zhiyong Xu and
                  Yuan Zhou and
                  Qinglin Lu and
                  Xuming He},
  title        = {Pack and Force Your Memory: Long-form and Consistent Video Generation},
  journal      = {ArXiv preprint},
  year         = {2025},
}

@article{memoryforcing,
  author       = {Junchao Huang and
                  Xinting Hu and
                  Boyao Han and
                  Shaoshuai Shi and
                  Zhuotao Tian and
                  Tianyu He and
                  Li Jiang},
  title        = {Memory Forcing: Spatio-Temporal Memory for Consistent Scene Generation
                  on Minecraft},
  journal      = {ArXiv preprint},
  year         = {2025},
}

@article{moc,
  author       = {Shengqu Cai and
                  Ceyuan Yang and
                  Lvmin Zhang and
                  Yuwei Guo and
                  Junfei Xiao and
                  Ziyan Yang and
                  Yinghao Xu and
                  Zhenheng Yang and
                  Alan L. Yuille and
                  Leonidas J. Guibas and
                  Maneesh Agrawala and
                  Lu Jiang and
                  Gordon Wetzstein},
  title        = {Mixture of Contexts for Long Video Generation},
  journal      = {ArXiv preprint},
  year         = {2025},
}

@inproceedings{tttvideo,
  author       = {Karan Dalal and
                  Daniel Koceja and
                  Jiarui Xu and
                  Yue Zhao and
                  Shihao Han and
                  Ka Chun Cheung and
                  Jan Kautz and
                  Yejin Choi and
                  Yu Sun and
                  Xiaolong Wang},
  title        = {One-Minute Video Generation with Test-Time Training},
  booktitle    = CVPR,
  year         = {2025},
}

@article{lact,
  author       = {Tianyuan Zhang and
                  Sai Bi and
                  Yicong Hong and
                  Kai Zhang and
                  Fujun Luan and
                  Songlin Yang and
                  Kalyan Sunkavalli and
                  William T. Freeman and
                  Hao Tan},
  title        = {Test-Time Training Done Right},
  journal      = {ArXiv preprint},
  year         = {2025},
}

@inproceedings{pfp,
  title     = {TinyHistory: Lightweight Video History Embeddings via Two-Stage Context Learning},
  author    = {Lvmin Zhang and Shengqu Cai and Muyang Li and Chong Zeng and Beijia Lu and Anyi Rao and Song Han and Gordon Wetzstein and Maneesh Agrawala},
  booktitle = ECCV,
  year      = {2026},
}

@article{longcat,
      title={LongCat-Video Technical Report}, 
      author={{Meituan LongCat Team}},
      year={2025},
  journal      = {ArXiv preprint},

}

@article{tf1,
  author       = {Yu Lu and
                  Yi Yang},
  title        = {FreeLong++: Training-Free Long Video Generation via Multi-band SpectralFusion},
  journal      = {ArXiv preprint},
  year         = {2025},
}

@inproceedings{tf2,
  author       = {Yu Lu and
                  Yuanzhi Liang and
                  Linchao Zhu and
                  Yi Yang},
  title        = {FreeLong: Training-Free Long Video Generation with SpectralBlend Temporal
                  Attention},
  booktitle    = NeurIPS,
  year         = {2024},
}

@inproceedings{diffusionforcing,
  author       = {Boyuan Chen and
                  Diego Marti Monso and
                  Yilun Du and
                  Max Simchowitz and
                  Russ Tedrake and
                  Vincent Sitzmann},
  title        = {Diffusion Forcing: Next-token Prediction Meets Full-Sequence Diffusion},
  booktitle    = NeurIPS,
  year         = {2024},
}

@inproceedings{hsg,
  author       = {Kiwhan Song and
                  Boyuan Chen and
                  Max Simchowitz and
                  Yilun Du and
                  Russ Tedrake and
                  Vincent Sitzmann},
  title        = {History-Guided Video Diffusion},
  booktitle    = ICML,
  year         = {2025},
}

@inproceedings{st2v,
  author       = {Roberto Henschel and
                  Levon Khachatryan and
                  Hayk Poghosyan and
                  Daniil Hayrapetyan and
                  Vahram Tadevosyan and
                  Zhangyang Wang and
                  Shant Navasardyan and
                  Humphrey Shi},
  title        = {StreamingT2V: Consistent, Dynamic, and Extendable Long Video Generation
                  from Text},
  booktitle    = CVPR,
  year         = {2025},
}

@article{far,
  author       = {Yuchao Gu and
                  Weijia Mao and
                  Mike Zheng Shou},
  title        = {Long-Context Autoregressive Video Modeling with Next-Frame Prediction},
  journal      = {ArXiv preprint},
  year         = {2025},
}

@article{sana-video,
  author       = {Junsong Chen and
                  Yuyang Zhao and
                  Jincheng Yu and
                  Ruihang Chu and
                  Junyu Chen and
                  Shuai Yang and
                  Xianbang Wang and
                  Yicheng Pan and
                  Daquan Zhou and
                  Huan Ling and
                  Haozhe Liu and
                  Hongwei Yi and
                  Hao Zhang and
                  Muyang Li and
                  Yukang Chen and
                  Han Cai and
                  Sanja Fidler and
                  Ping Luo and
                  Song Han and
                  Enze Xie},
  title        = {SANA-Video: Efficient Video Generation with Block Linear Diffusion
                  Transformer},
  journal      = {ArXiv preprint},
  year         = {2025},
}

@article{streamdit,
  author       = {Akio Kodaira and
                  Tingbo Hou and
                  Ji Hou and
                  Masayoshi Tomizuka and
                  Yue Zhao},
  title        = {StreamDiT: Real-Time Streaming Text-to-Video Generation},
  journal      = {ArXiv preprint},
  year         = {2025},
}

@inproceedings{po2025long,
  title={Long-context state-space video world models},
  author={Po, Ryan and Nitzan, Yotam and Zhang, Richard and Chen, Berlin and Dao, Tri and Shechtman, Eli and Wetzstein, Gordon and Huang, Xun},
  booktitle=ICCV,
  year={2025}
}

@article{LTXV,
  title={LTX-Video: Realtime Video Latent Diffusion},
  author={HaCohen, Yoav and Chiprut, Nisan and Brazowski, Benny and Shalem, Daniel and Moshe, Dudu and Richardson, Eitan and Levin, Eran and Shiran, Guy and Zabari, Nir and Gordon, Ori and Panet, Poriya and Weissbuch, Sapir and Kulikov, Victor and Bitterman, Yaki and Melumian, Zeev and Bibi, Ofir},
  journal      = {ArXiv preprint},
  year={2024}
}

@inproceedings{DC-AE,
  author       = {Junyu Chen and
                  Han Cai and
                  Junsong Chen and
                  Enze Xie and
                  Shang Yang and
                  Haotian Tang and
                  Muyang Li and
                  Song Han},
  title        = {Deep Compression Autoencoder for Efficient High-Resolution Diffusion
                  Models},
  booktitle    = ICLR,
  year         = {2025},
}

@inproceedings{dao2022flashattention,
  title={Flash{A}ttention: Fast and Memory-Efficient Exact Attention with {IO}-Awareness},
  author={Dao, Tri and Fu, Daniel Y. and Ermon, Stefano and Rudra, Atri and R{\'e}, Christopher},
  booktitle=NeurIPS,
  year={2022}
}

@inproceedings{dao2023flashattention2,
  title={Flash{A}ttention-2: Faster Attention with Better Parallelism and Work Partitioning},
  author={Dao, Tri},
  booktitle=ICLR,
  year={2024}
}

@inproceedings{zhang2025sageattention,
  title={SageAttention: Accurate 8-Bit Attention for Plug-and-play Inference Acceleration}, 
  author={Zhang, Jintao and Wei, Jia and Zhang, Pengle and Zhu, Jun and Chen, Jianfei},
  booktitle=ICLR,
  year={2025}
}

@inproceedings{zhang2024sageattention2,
  title={Sageattention2: Efficient attention with thorough outlier smoothing and per-thread int4 quantization},
  author={Zhang, Jintao and Huang, Haofeng and Zhang, Pengle and Wei, Jia and Zhu, Jun and Chen, Jianfei},
  booktitle=ICML,
  year={2025}
}

@article{zhang2025sageattention3,
  title={SageAttention3: Microscaling FP4 Attention for Inference and An Exploration of 8-Bit Training},
  author={Zhang, Jintao and Wei, Jia and Zhang, Pengle and Xu, Xiaoming and Huang, Haofeng and Wang, Haoxu and Jiang, Kai and Zhu, Jun and Chen, Jianfei},
  journal      = {ArXiv preprint},
  year={2025}
}

@inproceedings{radial-attn,
  author       = {Xingyang Li and
                  Muyang Li and
                  Tianle Cai and
                  Haocheng Xi and
                  Shuo Yang and
                  Yujun Lin and
                  Lvmin Zhang and
                  Songlin Yang and
                  Jinbo Hu and
                  Kelly Peng and
                  Maneesh Agrawala and
                  Ion Stoica and
                  Kurt Keutzer and
                  Song Han},
  title        = {Radial Attention: O(n log n) Sparse Attention with Energy Decay for
                  Long Video Generation},
  booktitle=NeurIPS,
  year         = {2025},
}

@inproceedings{xue2025ultravideo,
  title={UltraVideo: High-Quality UHD Video Dataset with Comprehensive Captions},
  author={Xue, Zhucun and Zhang, Jiangning and Hu, Teng and He, Haoyang and Chen, Yinan and Cai, Yuxuan and Wang, Yabiao and Wang, Chengjie and Liu, Yong and Li, Xiangtai and others},
  booktitle=NeurIPS,
  year={2025}
}

@inproceedings{XCLIP,
  title     = {{X-CLIP}: End-to-End Multi-grained Contrastive Learning for Video-Text Retrieval},
  author    = {Yiwei Ma and Guohai Xu and Xiaoshuai Sun and Ming Yan and Ji Zhang and Rongrong Ji},
  booktitle = ACMMM,
  year      = {2022},
  doi       = {10.1145/3503161.3547910},
}

@inproceedings{liu2023flow,
  title={Flow Straight and Fast: Learning to Generate and Transfer Data with Rectified Flow},
  author={Xingchao Liu and Chengyue Gong and Qiang Liu},
  booktitle=ICLR,
  year={2023}
}

@article{yu2025videossm,
  title={VideoSSM: Autoregressive Long Video Generation with Hybrid State-Space Memory},
  author={Yu, Yifei and Wu, Xiaoshan and Hu, Xinting and Hu, Tao and Sun, Yangtian and Lyu, Xiaoyang and Wang, Bo and Ma, Lin and Ma, Yuewen and Wang, Zhongrui and others},
  journal={ArXiv preprint},
  year={2025}
}

@inproceedings{esser2024scaling,
	title={Scaling rectified flow transformers for high-resolution image synthesis},
	author={Esser, Patrick and Kulal, Sumith and Blattmann, Andreas and Entezari, Rahim and M{\"u}ller, Jonas and Saini, Harry and Levi, Yam and Lorenz, Dominik and Sauer, Axel and Boesel, Frederic and others},
	booktitle=ICML,
	year={2024}
}

@inproceedings{zhang2018unreasonable,
  title={The unreasonable effectiveness of deep features as a perceptual metric},
  author={Zhang, Richard and Isola, Phillip and Efros, Alexei A and Shechtman, Eli and Wang, Oliver},
  booktitle=CVPR,
  year={2018}
}

@misc{schuhmann2022improved,
  author       = {Schuhmann, Christoph},
  title        = {Improved Aesthetic Predictor},
  year         = {2022},
  howpublished = {\url{https://github.com/christophschuhmann/improved-aesthetic-predictor}},
  note         = {Accessed: 2026-06-30},
}

@article{Qwen3-VL,
  title   = {Qwen3-VL Technical Report},
  author  = {{Qwen Team}},
  journal = {ArXiv preprint},
  year    = {2025},
  url     = {https://arxiv.org/abs/2511.21631},
}

@article{carion2025sam,
  title={Sam 3: Segment anything with concepts},
  author={Carion, Nicolas and Gustafson, Laura and Hu, Yuan-Ting and Debnath, Shoubhik and Hu, Ronghang and Suris, Didac and Ryali, Chaitanya and Alwala, Kalyan Vasudev and Khedr, Haitham and Huang, Andrew and others},
  journal={ArXiv preprint},
  year={2025}
}

@inproceedings{sun2021loftr,
  title={LoFTR: Detector-free local feature matching with transformers},
  author={Sun, Jiaming and Shen, Zehong and Wang, Yuang and Bao, Hujun and Zhou, Xiaowei},
  booktitle=CVPR,
  year={2021}
}

@inproceedings{hu2022lora,
  title     = {{LoRA}: Low-Rank Adaptation of Large Language Models},
  author    = {Hu, Edward J. and Shen, Yelong and Wallis, Phillip and Allen-Zhu, Zeyuan and Li, Yuanzhi and Wang, Shean and Wang, Lu and Chen, Weizhu},
  booktitle = ICLR,
  year      = {2022},
  url       = {https://openreview.net/forum?id=nZeVKeeFYf9},
}

@article{simeoni2025dinov3,
  title={Dinov3},
  author={Sim{\'e}oni, Oriane and Vo, Huy V and Seitzer, Maximilian and Baldassarre, Federico and Oquab, Maxime and Jose, Cijo and Khalidov, Vasil and Szafraniec, Marc and Yi, Seungeun and Ramamonjisoa, Micha{\"e}l and others},
  journal={ArXiv preprint},
  year={2025}
}

@article{lowe2004SIFT,
  title={Distinctive image features from scale-invariant keypoints},
  author={Lowe, David G},
  journal={International journal of computer vision},
  year={2004},
}

@inproceedings{AMT,
  author       = {Zhen Li and
                  Zuo{-}Liang Zhu and
                  Linghao Han and
                  Qibin Hou and
                  Chun{-}Le Guo and
                  Ming{-}Ming Cheng},
  title        = {{AMT:} All-Pairs Multi-Field Transforms for Efficient Frame Interpolation},
  booktitle    = CVPR,
  year         = {2023},
}

@article{hong2025relic,
  title        = {{RELIC}: Interactive Video World Model with Long-Horizon Memory},
  author       = {Hong, Yicong and Mei, Yiqun and Ge, Chongjian and Xu, Yiran and Zhou, Yang and Bi, Sai and Hold{-}Geoffroy, Yannick and Roberts, Mike and Fisher, Matthew and Shechtman, Eli and Sunkavalli, Kalyan and Liu, Feng and Li, Zhengqi and Tan, Hao},
  journal      = {ArXiv preprint},
  year         = {2025}
}

@article{oshima2025worldpack,
  title        = {{WorldPack}: Compressed Memory Improves Spatial Consistency in Video World Modeling},
  author       = {Oshima, Yuta and Iwasawa, Yusuke and Suzuki, Masahiro and Matsuo, Yutaka and Furuta, Hiroki},
  journal      = {ArXiv preprint},
  year         = {2025}
}

\appendix
\renewcommand{\thefigure}{A\arabic{figure}}
\renewcommand{\thetable}{A\arabic{table}}
\newpage
\section{Experimental Settings} \label{apx:settings}
\subsubsection{Ring-Structured Data Construction Details}

This section supplements the mathematical implementation details and hyperparameter settings for the ring-structured sequence construction introduced in Section 3.1 of the main paper.

\paragraph{Random Head Cropping} As described in the main text, to prevent information leakage, we apply a random cropping strategy to the head of the constructed full context $\mathbf{C}_{full}$. Specifically, let $s$ denote the starting index of the reversed target clip $\mathbf{x}_{tgt}^{rev}$ within the synthetic history $\mathbf{h}_{syn}$. To ensure the shortcut is removed without discarding the embedded ``answer'', we sample a crop length $l \sim \text{Uniform}(0, s)$ and strictly remove the first $l$ frames from $\mathbf{C}_{full}$, yielding the cropped context $\mathbf{C}_{crop}$.

\paragraph{Ring Drop Probability} In Equation (3) of the main text, we introduced the Bernoulli variable $b \sim \text{Bernoulli}(1 - p_{\text{drop}})$ to balance the model's historical adherence against open-ended generation diversity. Across all main experiments in this paper, unless otherwise specified in the ablation studies, we set the default ring context drop probability to $p_{\text{drop}} = 0.4$.

\subsubsection{Compression and Timestep Composition}

To efficiently process minutes-long history without quadratic computational explosion, we adopt a dual-stream composition strategy based on the diffusion timestep $t$. In Wan models, the timestep $t$ represents the noise level, defined continuously within $t \in [0, 1]$, where $t=1$ denotes pure noise and $t=0$ denotes the clean data distribution. 

To perfectly align with the intrinsic noise schedule and pre-trained timestep boundaries of the base model, we specifically set the transition threshold to $\tau = 0.417$, which corresponds to the native pre-training configuration of Wan2.2-A14B. 
Specifically, during the high-noise phase ($t \in [0.417, 1]$), the model determines the global semantic structure and motion dynamics; thus, we apply the global stream with a compression configuration of $\Psi(Z_{hist}; 2, 2, 8)$ to retain temporal density. Conversely, during the low-noise phase ($t \in [0, 0.417]$), the model focuses on high-frequency texture and local detail refinement. In this stage, we switch to the detail stream $\Psi(Z_{hist}; 1, 1, 32)$ and apply a period-32 cyclic shift mechanism. At each iteration, we sample an offset $\delta \in \{0, \dots, 31\}$ and select temporal indices $\mathcal{I}_{\delta} = \{i \mid i \equiv \delta \pmod{32}\}$ to ensure complete coverage of the uncompressed historical details across different sampling steps.

\subsubsection{Training Hyperparameters}

Both the Wan2.1-1.3B and Wan2.2-A14B models are fine-tuned using Low-Rank Adaptation (LoRA). Specifically, with the LoRA rank set to $r=128$, the trainable parameters are strategically injected into the attention layers and Feed-Forward Networks (FFNs) within the Diffusion Transformer (DiT) blocks. 

The training is distributed across 4 compute nodes, each equipped with 8 NVIDIA H800 GPUs, utilizing a total of 32 GPUs via Accelerate and DeepSpeed/FSDP configurations. We optimize the models using the AdamW optimizer with a base learning rate of $1 \times 10^{-4}$. To ensure sufficient exposure to the long-video samples while maintaining training efficiency, the optimization is conducted for 7,000 steps, which takes approximately 30 hours in total.

\subsubsection{Computational Overhead}

Despite integrating an exceptionally long historical context, Ring Forcing maintains strict computational efficiency. By employing the compression operator $\Psi$ such that the product of the downsampling factors is exactly 32 (e.g., $2 \times 2 \times 8 = 32$ or $1 \times 1 \times 32 = 32$), the sequence length of the historical context is rigorously constrained. Our total sequence length matches the standard token budget $\mathcal{B}_{max} \approx 32,760$ of the original Wan model under the 5.4-second generation setting ($480 \times 832$ resolution, 81 frames). 

In our setup, we generate 1-second video clips at each autoregressive step. While a standard uncompressed history under this token budget would only span 4.4 seconds, applying our $32\times$ compression effectively extends the equivalent original history length to 140.8 seconds. Consequently, during the inference phase, a single generation step of our method---even when conditioned on 140.8 seconds of historical context---incurs the exact same computational cost and VRAM footprint as the original Wan model. Thus, it can be seamlessly deployed on any hardware capable of running the base Wan models.

\section{Limitations and Future Work}\label{apx:limitation}

While Ring Forcing significantly improves long-term memory and establishes robust object permanence in autoregressive video diffusion, minutes-long generation remains a challenging open problem. First, because the rollout is inherently causal and autoregressive, small inaccuracies in appearance, motion, or scene details can inevitably compound over extremely long horizons. Although our method drastically suppresses identity drift, subtle local distortions may still emerge across extended temporal spans.

Second, our training signal is derived from the ring-based data construction, which serves as a highly efficient proxy task to enforce long-range retrieval. However, extreme real-world scenarios---such as multiple visually similar objects undergoing dense, intersecting occlusions---demand even more fine-grained memory extraction capabilities. This highlights a promising direction for future work to enhance retrieval mechanisms in multi-object and complex interactive environments.

Finally, we currently expand memory capacity through a fixed-rate compression and timestep-dependent composition strategy. While this hard constraint successfully bounds the VRAM footprint to be strictly identical to the base model, it lacks flexibility. Fixed spatiotemporal downsampling inevitably discards some high-frequency information, which can occasionally affect extremely small objects or fast transient motions. A critical and inspiring avenue for future work is to explore flexible, \textbf{content-adaptive memory compression mechanisms}. Such mechanisms would dynamically allocate token budgets based on spatial complexity (e.g., preserving high resolution for critical subjects while aggressively compressing static backgrounds) and adaptively adjust compression ratios to accommodate \textbf{varying computational budgets}. This will enable a more elegant and scalable trade-off between generative fidelity and hardware resource utilization. Future efforts will also include extending evaluation protocols to minute-scale, multi-shot scenarios.

\section{Additional Results}\label{apx:vis}

In this section, we provide extended qualitative results to further examine the long-term memory and general video generation capabilities of Ring Forcing. 

First, we present additional comparisons on the Appear-Disappear-Reappear (A-D-R) benchmark under 1-second and 5-second occlusion gaps. These examples visually illustrate the prevalent issue of identity drift in existing autoregressive baselines and highlight our method's capacity to maintain strict object permanence even when the target subject temporarily exits the camera frustum.

Second, we push the temporal boundaries by evaluating our model's performance under extreme 15-second, 30-second, and 60-second disappearance gaps. These stress-test results demonstrate that Ring Forcing can reliably retrieve specific fine-grained details from distant history, effectively bridging prolonged visual disconnects without relying on short-term attention biases.

Finally, we include broader examples of 60-second continuous video generation in unconstrained, general scenarios. These long-horizon rollouts verify that the proposed memory mechanisms do not compromise the base model's inherent generative priors, consistently yielding videos with high aesthetic quality, natural dynamics, and coherent spatiotemporal progression.
\begin{figure}[htbp]
    \centering
    \includegraphics[width=\textwidth]{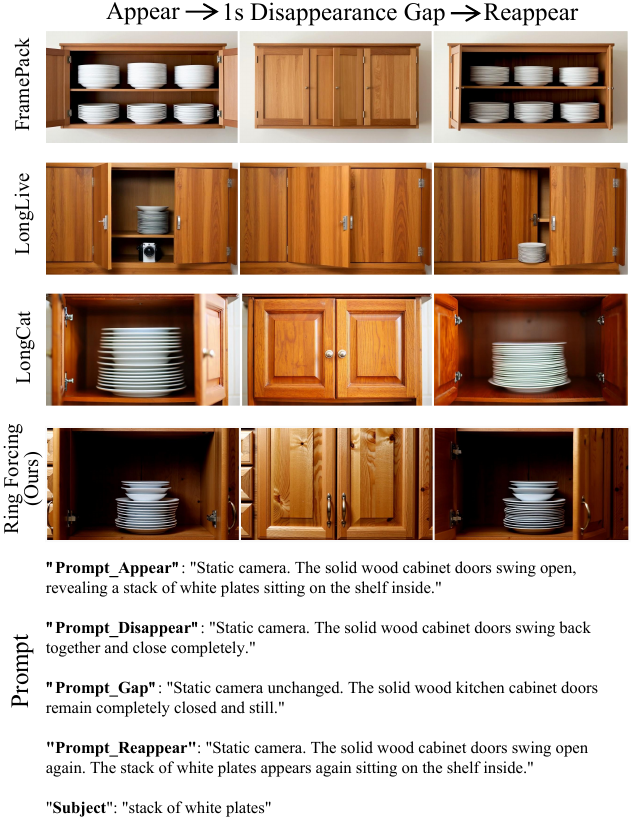}
    \caption{\textbf{Qualitative Comparison on the A-D-R Benchmark (1s Gap).} While state-of-the-art baselines suffer from immediate identity drift and fail to reconstruct the hidden object (stack of white plates), Ring Forcing accurately retrieves historical information, ensuring strict object permanence.}
    \label{fig:supp_1s_plates}
\end{figure}

\begin{figure}[htbp]
    \centering
    \includegraphics[width=\textwidth]{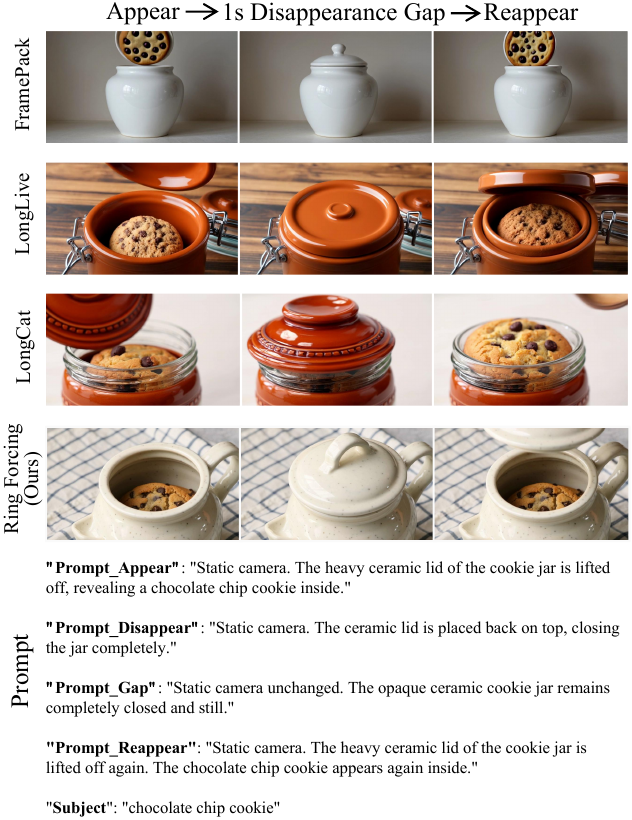}
    \caption{\textbf{Qualitative Comparison on the A-D-R Benchmark (1s Gap).} Existing autoregressive models exhibit myopic attention bias, altering the geometry and texture of the subject (cookie) after merely one second of occlusion. In contrast, our method maintains strong attribute retention.}
    \label{fig:supp_1s_cookie}
\end{figure}

\begin{figure}[htbp]
    \centering
    \includegraphics[width=\textwidth]{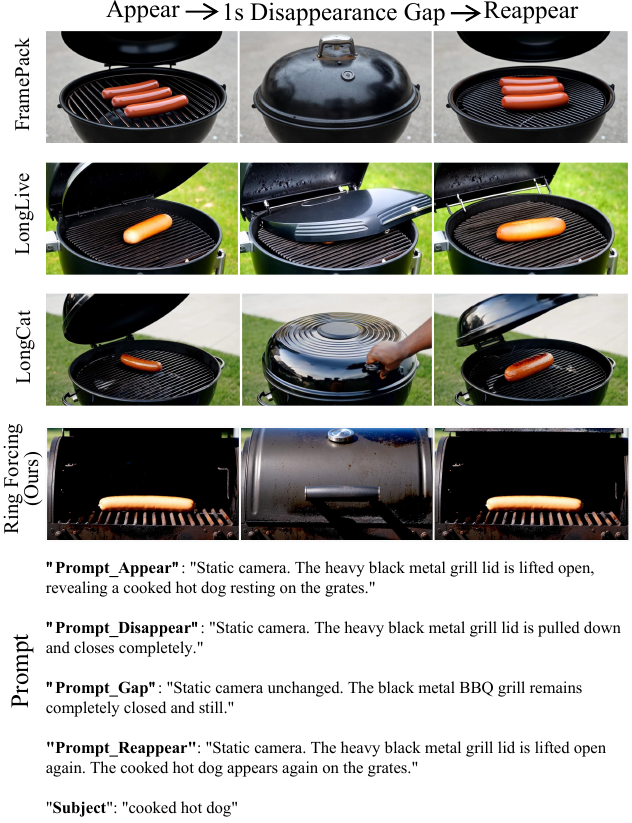}
    \caption{\textbf{Qualitative Comparison on the A-D-R Benchmark (1s Gap).} Even with simple objects (hot dog), baseline models struggle to maintain consistency. Ring Forcing effectively bridges the temporal gap, accurately recovering the subject's identity.}
    \label{fig:supp_1s_hotdog}
\end{figure}
\begin{figure}[h
tbp]
    \centering
    \includegraphics[width=\textwidth]{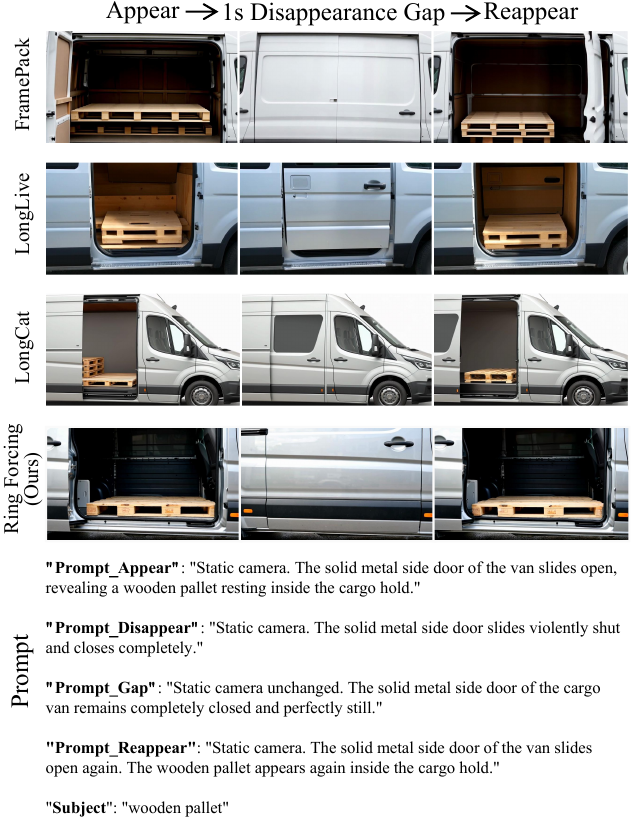}
    \caption{\textbf{Qualitative Comparison on the A-D-R Benchmark (1s Gap).} While baseline models struggle to maintain the structural and textural details of the hidden object (wooden pallet) after a brief occlusion, Ring Forcing successfully retrieves the precise historical information, ensuring strict object permanence.}
    \label{fig:supp_1s_hotdog}
\end{figure}
\begin{figure}[htbp]
    \centering
    \includegraphics[width=\textwidth]{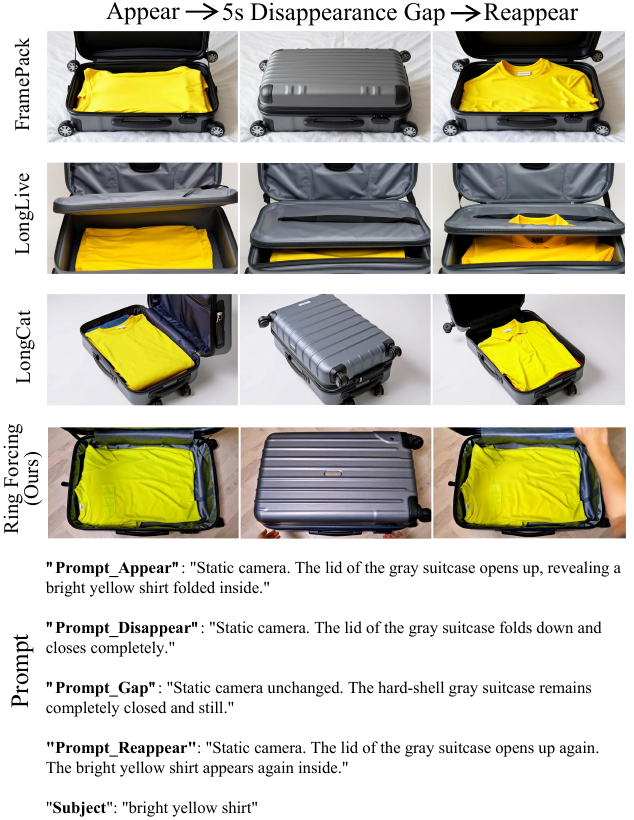}
    \caption{\textbf{Qualitative Comparison on the A-D-R Benchmark (5s Gap).} As the temporal divide expands, baseline models experience catastrophic forgetting. Ring Forcing reliably preserves the specific visual features of the subject (yellow shirt) from distant history.}
    \label{fig:supp_5s_shirt}
\end{figure}

\begin{figure}[htbp]
    \centering
    \includegraphics[width=\textwidth]{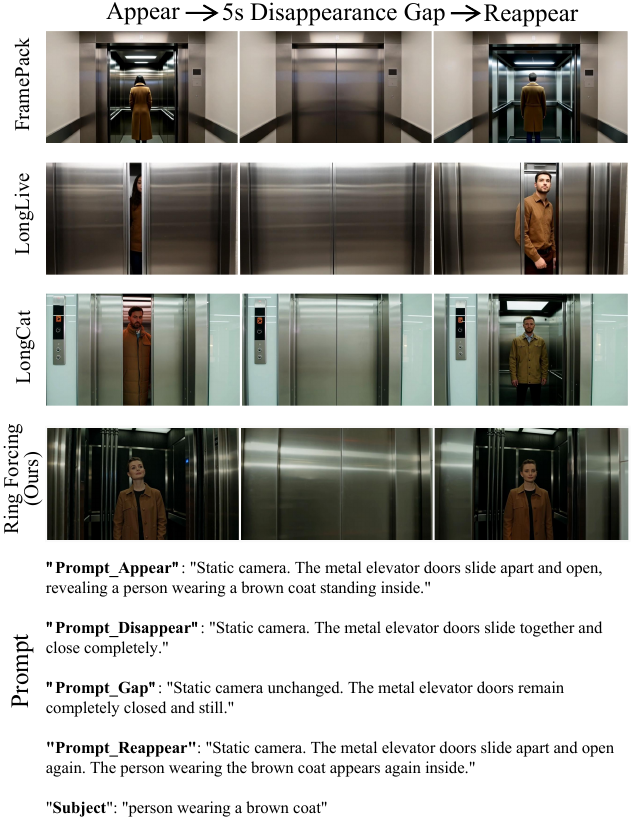}
    \caption{\textbf{Qualitative Comparison on the A-D-R Benchmark (5s Gap).} Our method successfully reconstructs complex human attributes across a 5-second occlusion, whereas baselines hallucinate entirely new identities, highlighting their inability to perform long-range retrieval.}
    \label{fig:supp_5s_person}
\end{figure}

\begin{figure}[htbp]
    \centering
    \includegraphics[width=\textwidth]{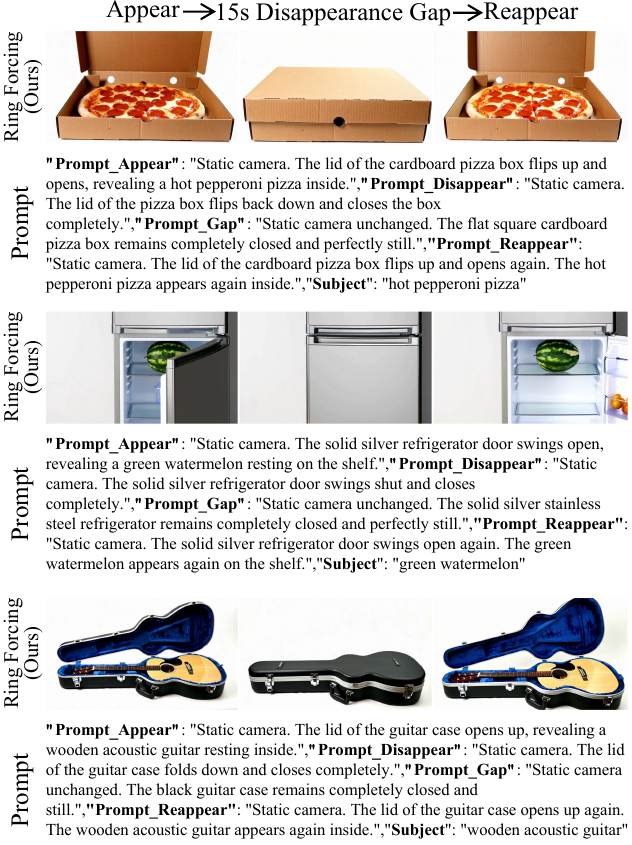}
    \caption{\textbf{Extreme Long-Term Memory Retrieval (15s Gap).} Operating well beyond the maximum context limits of current baselines, Ring Forcing accurately recovers diverse objects (pizza, watermelon, acoustic guitar) after a 15-second disappearance, demonstrating highly resilient memory capacity.}
    \label{fig:supp_15s_extreme}
\end{figure}

\begin{figure}[htbp]
    \centering
    \includegraphics[width=\textwidth]{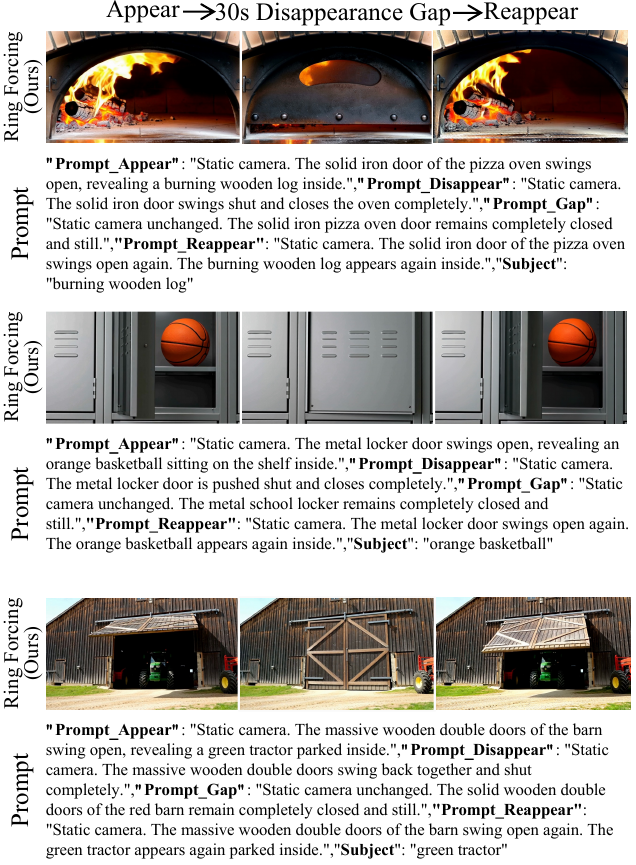}
    \caption{\textbf{Extreme Long-Term Memory Retrieval (30s Gap).} Stress-testing our method with a half-minute visual gap. Ring Forcing exhibits robust temporal consistency and semantic adherence even for dynamic scenes (burning log, basketball, tractor), successfully overcoming the bottleneck of prolonged occlusions.}
    \label{fig:supp_30s_extreme}
\end{figure}

\begin{figure}[htbp]
    \centering
    \includegraphics[width=\textwidth]{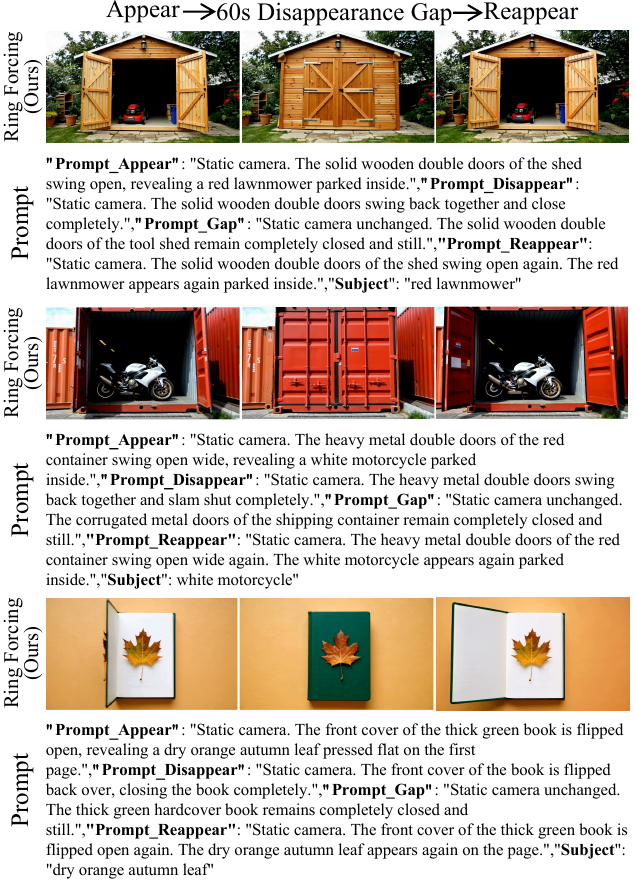}
    \caption{\textbf{Long-Term Object Permanence Evaluation (60s Gap).} Demonstrating minute-level memory preservation, Ring Forcing successfully retrieves precise high-frequency details (lawnmower, motorcycle, autumn leaf) across a 60-second temporal gap, further demonstrating the effectiveness of Ring Forcing.}
    \label{fig:supp_60s_extreme}
\end{figure}

\begin{figure*}[htbp!]
    \centering
    \includegraphics[width=0.85\linewidth]{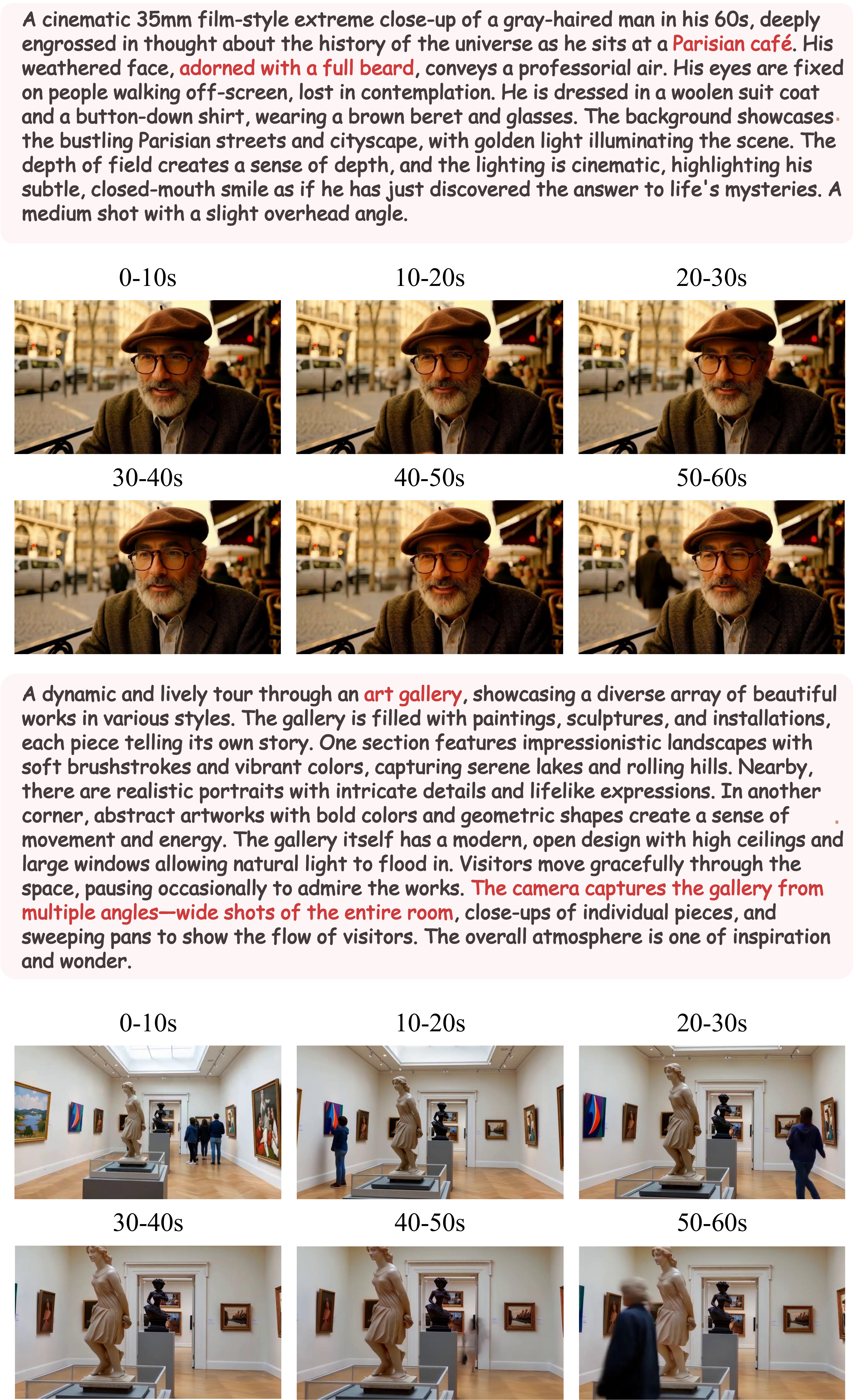}
    \caption{\textbf{Qualitative Results on General Long Video Generation (1/5).} This example demonstrates the model's capability to maintain visual quality, stable dynamics, and overall spatiotemporal consistency throughout a 1-minute continuous generation in general scenarios.}
    \label{fig:apx1}
\end{figure*}

\begin{figure*}[htbp!]
    \centering
    \includegraphics[width=0.85\linewidth]{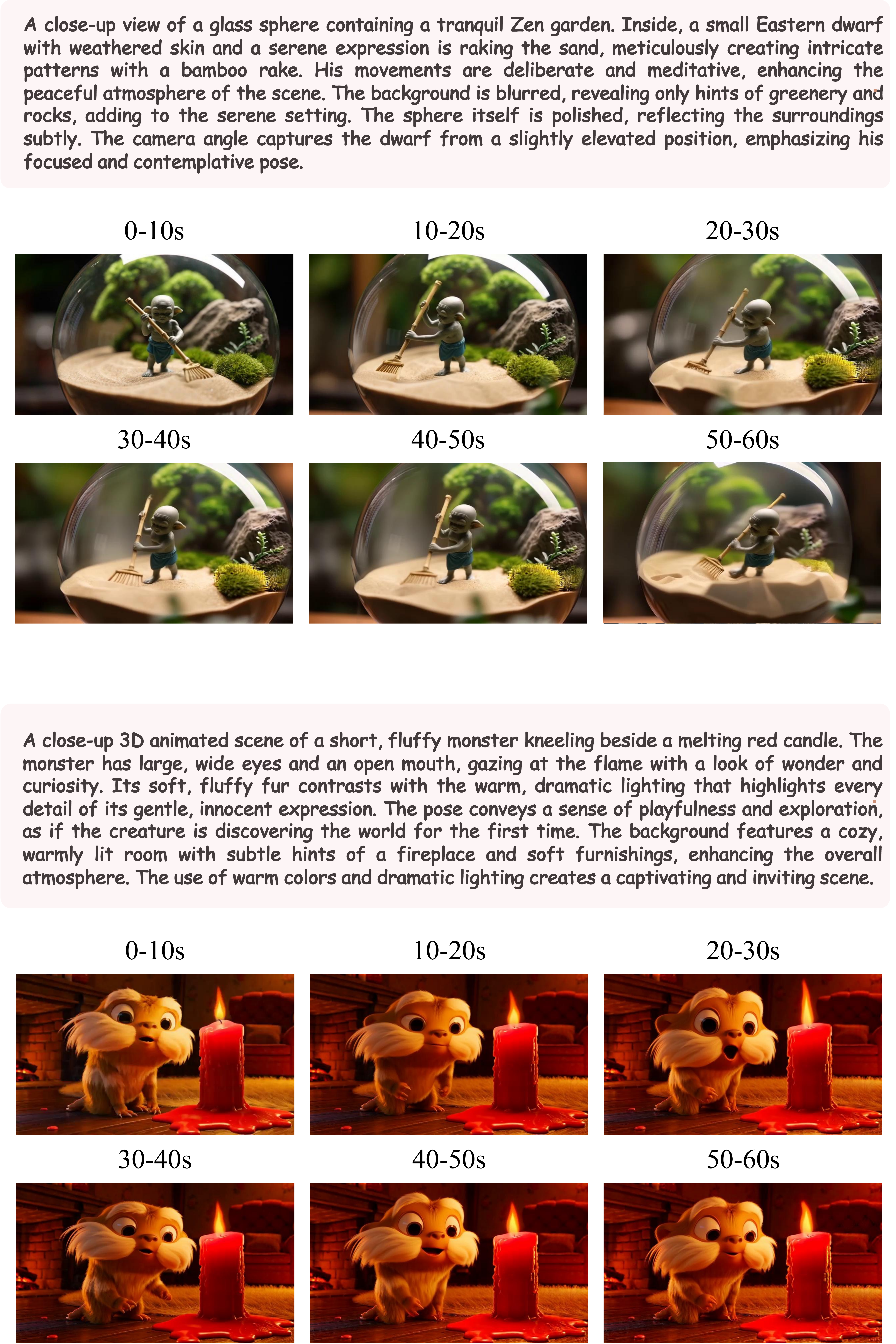}
    \caption{\textbf{Qualitative Results on General Long Video Generation (2/5).} This example demonstrates the model's capability to maintain visual quality, stable dynamics, and overall spatiotemporal consistency throughout a 1-minute continuous generation in general scenarios.}
    \label{fig:apx2}
\end{figure*}

\begin{figure*}[htbp!]
    \centering
    \includegraphics[width=0.85\linewidth]{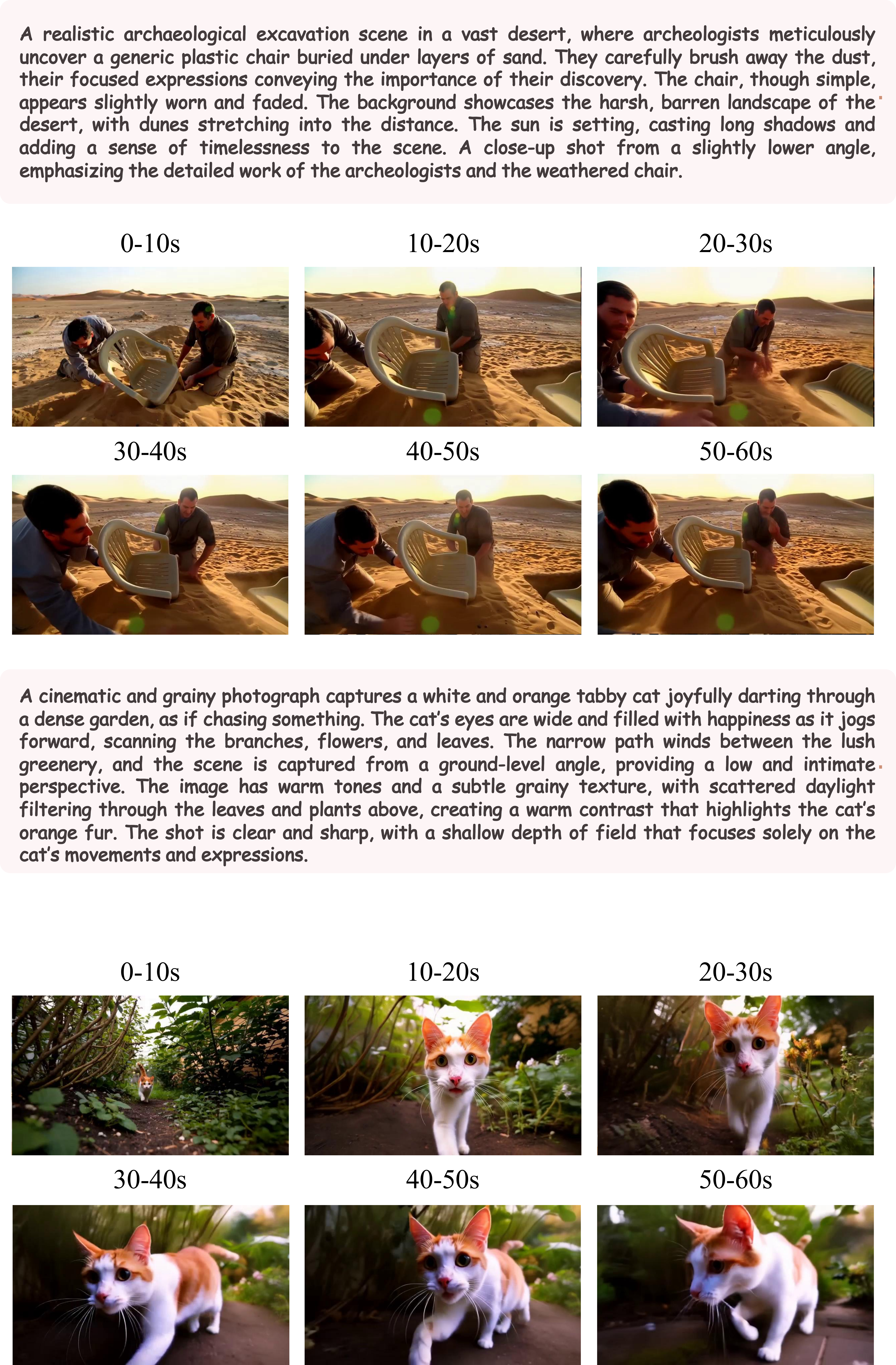}
    \caption{\textbf{Qualitative Results on General Long Video Generation (3/5).} This example demonstrates the model's capability to maintain visual quality, stable dynamics, and overall spatiotemporal consistency throughout a 1-minute continuous generation in general scenarios.}
    \label{fig:apx3}
\end{figure*}

\begin{figure*}[htbp!]
    \centering
    \includegraphics[width=0.85\linewidth]{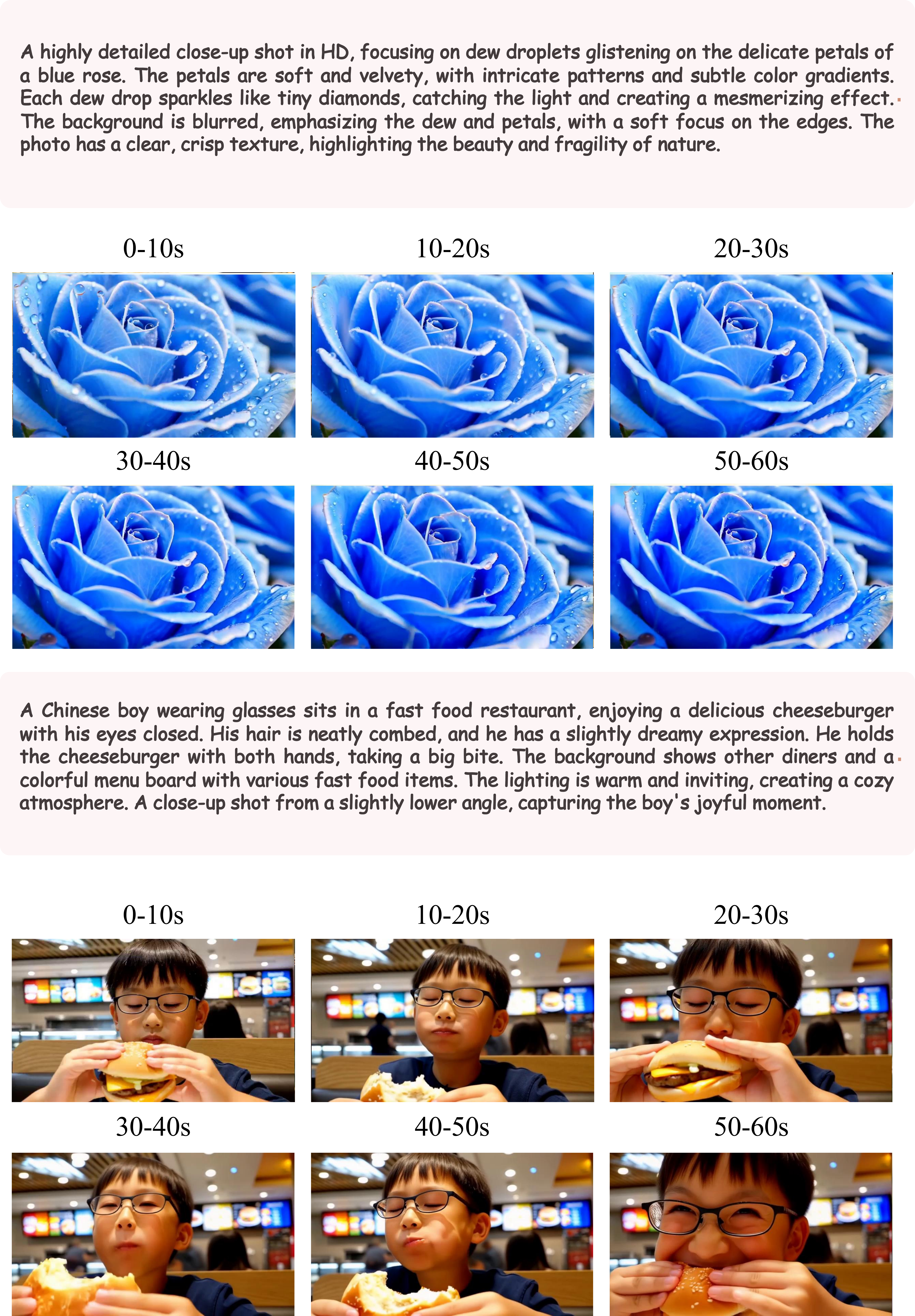}
    \caption{\textbf{Qualitative Results on General Long Video Generation (4/5).} This example demonstrates the model's capability to maintain visual quality, stable dynamics, and overall spatiotemporal consistency throughout a 1-minute continuous generation in general scenarios.}
    \label{fig:apx4}
\end{figure*}

\begin{figure*}[htbp!]
    \centering
    \includegraphics[width=0.85\linewidth]{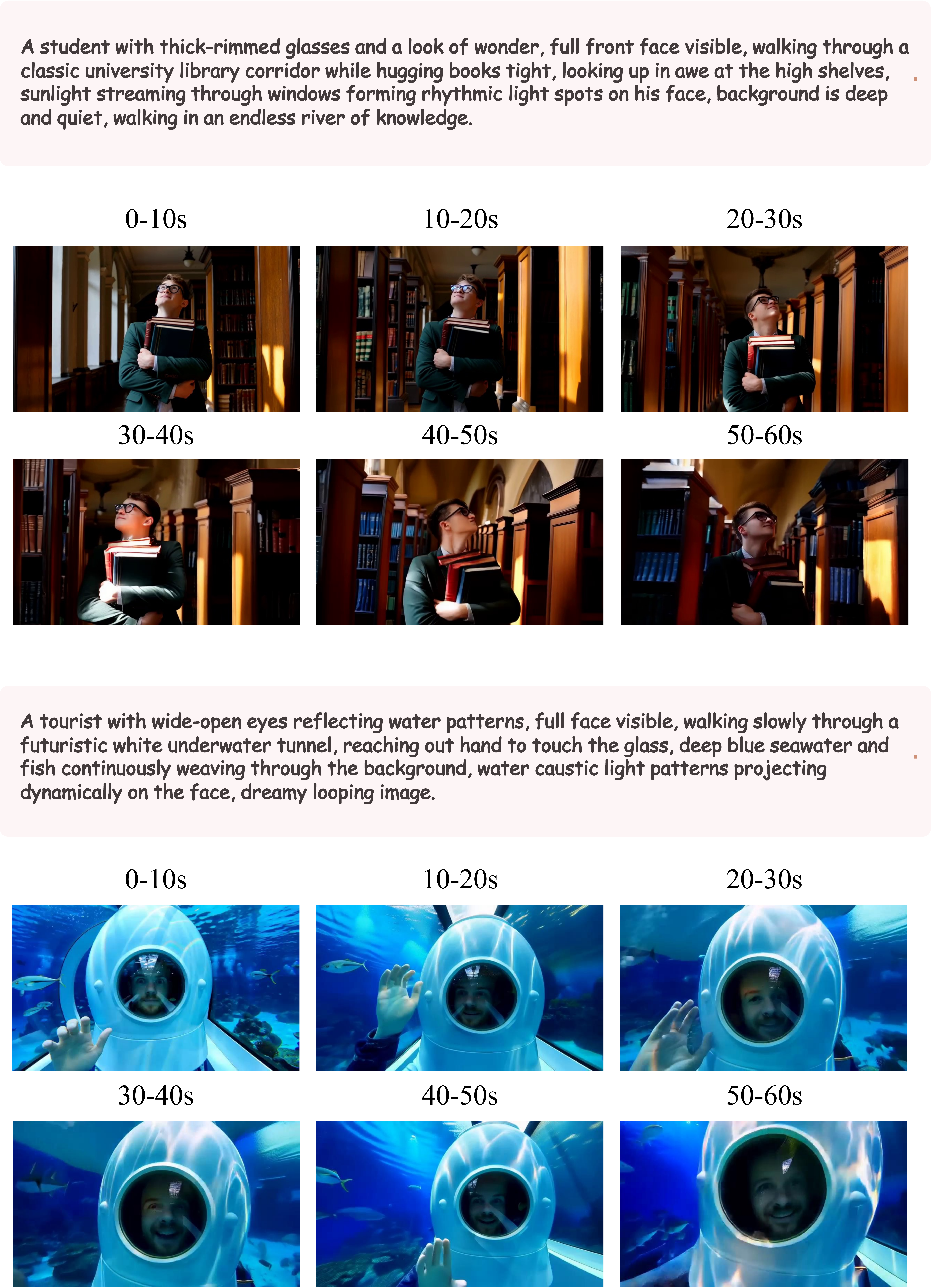}
    \caption{\textbf{Qualitative Results on General Long Video Generation (5/5).} This example demonstrates the model's capability to maintain visual quality, stable dynamics, and overall spatiotemporal consistency throughout a 1-minute continuous generation in general scenarios.}
    \label{fig:apx5}
\end{figure*}

\end{document}